\pdfoutput=1
\documentclass[11pt]{article}

\usepackage[final]{acl}

\usepackage{times}
\usepackage{latexsym}

\usepackage[T1]{fontenc}
\usepackage[utf8]{inputenc}

\usepackage{microtype}

\usepackage{inconsolata}

\usepackage{booktabs}
\usepackage{graphicx}
\usepackage{siunitx}
\usepackage{here}
\usepackage{tabularx}
\usepackage{amssymb}
\usepackage{multirow}
\usepackage{ctable}
\usepackage{makecell}
\usepackage[subrefformat=parens]{subcaption}
\usepackage{multirow}
\usepackage{textcomp}
\usepackage{amsmath}
\usepackage[subrefformat=parens,labelformat=parens,skip=1pt]{subcaption}
\newcommand\fnsep{\textsuperscript{,}}

\title{Exploring the Effects of Alignment on Numerical Bias \\ in Large Language Models}

\author{
Ayako Sato$^{1,2,3}$,
Hwichan Kim$^{1,2}$,
Zhousi Chen$^{1,2}$,
Masato Mita$^{1,2}$,
Mamoru Komachi$^{2}$ \\
$^1$Tokyo Metropolitan University, 
$^2$Hitotsubashi University, 
$^3$CyberAgent Inc. \\
\texttt{sato\_ayako\_xa@cyberagent.co.jp,}\\
\texttt{\{kim@edu.sds, zhousi.chen@r, mita@edu.sds, mamoru.komachi@r\}.hit-u.ac.jp}
}

\begin{document}
\maketitle
\begin{abstract}
``LLM-as-a-judge,'' which utilizes large language models (LLMs) as evaluators, has proven effective in many evaluation tasks.
However, evaluator LLMs exhibit numerical bias, a phenomenon where certain evaluation scores are generated disproportionately often, leading reduced evaluation performance.
This study investigates the cause of this bias.
Given that most evaluator LLMs are aligned through instruction tuning and preference tuning, and that prior research suggests alignment reduces output diversity, we hypothesize that numerical bias arises from alignment.
To test this, we compare outputs from pre- and post-alignment LLMs, and observe that alignment indeed increases numerical bias.
We also explore mitigation strategies for post-alignment LLMs, including temperature scaling, distribution calibration, and score range adjustment.
Among these, score range adjustment is most effective in reducing bias and improving performance, though still heuristic.
% Among these methods, our experimental results demonstrate that employing an appropriate score range effectively reduces numerical bias while simultaneously improving evaluation performance.
Our findings highlight the need for further work on optimal score range selection and more robust mitigation strategies.
\end{abstract}

\begin{figure}[t]
    \setlength\abovecaptionskip{2truemm}
    \centering
    \includegraphics[width=1\linewidth]{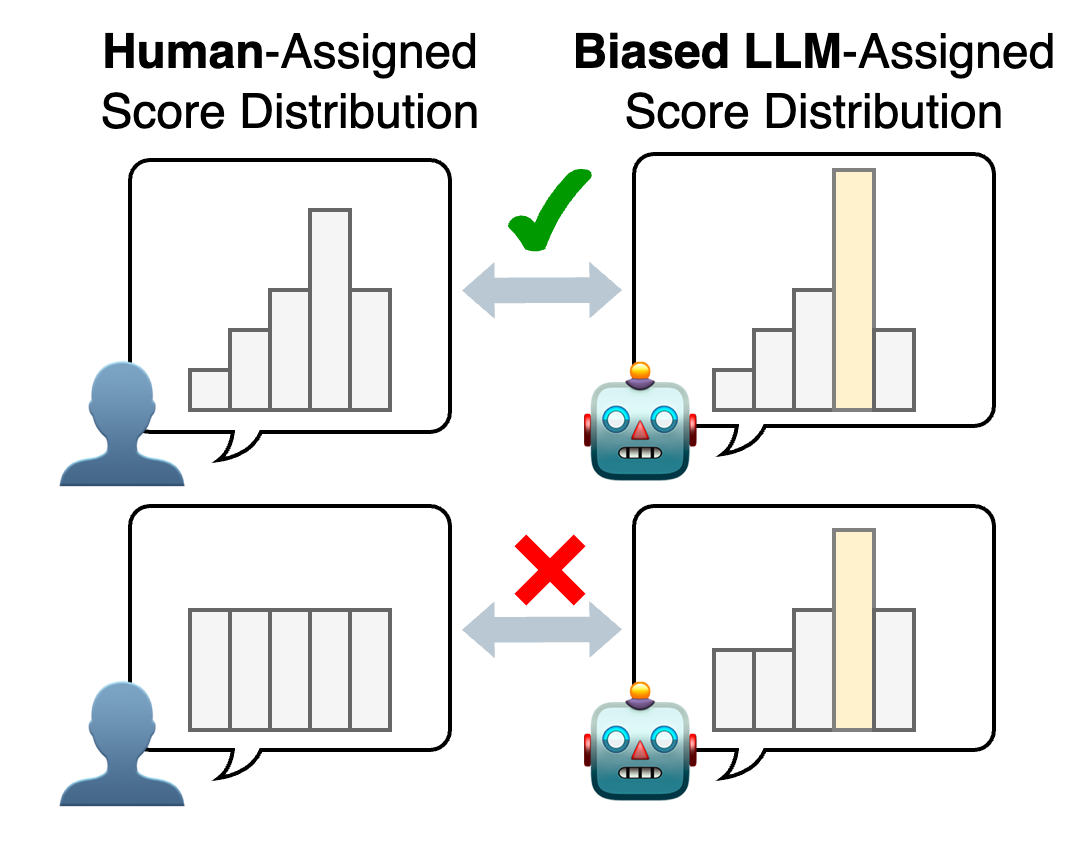}
    \caption{LLM evaluators with numerical biases tend to produce skewed scores regardless of input. In some cases, the results deviate from the distribution of manually assigned scores, questioning the robustness of the evaluation framework.}
    \label{fig:exapmle-bias}
\end{figure}

\section{Introduction}
As a novel approach for evaluating text quality, ``LLM-as-a-judge'' \cite{zheng2023judging}, which uses large language models (LLMs) as evaluators, has been gaining attention.
Traditional automatic evaluation metrics, such as BLEU \cite{papineni-etal-2002-bleu} and ROUGE \cite{lin-2004-rouge}, assess quality based on overlap with reference texts.
In contrast, LLM-as-a-judge directly evaluates input without references, generating scores or natural language outputs.
This approach reduces annotator workload and enables scalability.
Its effectiveness has been demonstrated across various evaluation tasks \cite{kocmi-federmann-2023-gemba, kocmi-federmann-2023-large, sato-etal-2024-tmu, kobayashi-etal-2024-large, enomoto-etal-2024-tmu}.
However, LLM-as-a-judge has certain challenges, such as \emph{numerical bias}—a tendency to overuse specific numeric tokens \cite{kocmi-federmann-2023-large, stureborg2024largelanguagemodelsinconsistent}.
An uneven distribution caused by numerical bias lacks \textit{distinguishability}, leading to higher overall correlation when matching the human evaluation distribution but raising robustness concerns as correlation may drop in case of a mismatch (Figure \ref{fig:exapmle-bias}).
Thus, identifying and mitigating numerical bias is crucial.

\begin{figure}[t]
    \centering
    \renewcommand{\arraystretch}{0.9}
    \begin{minipage}[b]{\linewidth}
        \centering
        \includegraphics[width=\linewidth]{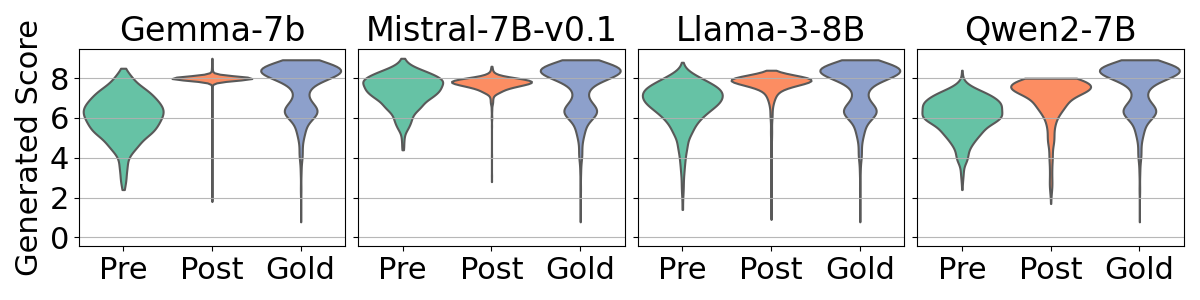}
        \subcaption{English to German (En-De)}
    \end{minipage}
    \begin{minipage}[b]{\linewidth}
        \centering
        \includegraphics[width=\linewidth]{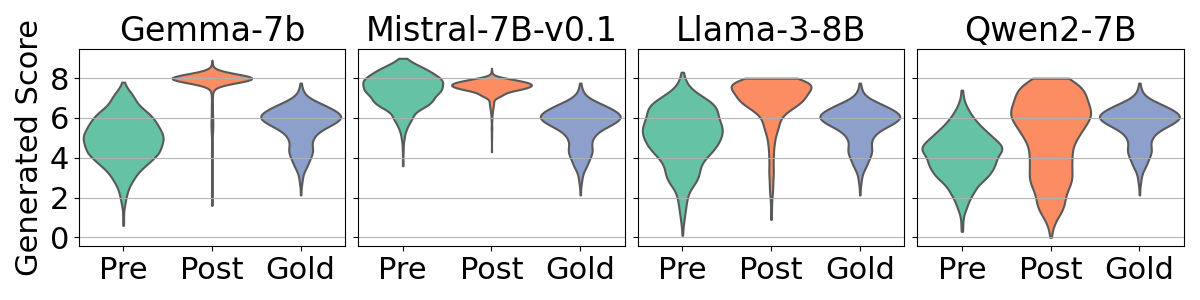}
        \subcaption{English to Chinese (En-Zh)}
    \end{minipage}
    \begin{minipage}[b]{\linewidth}
        \centering
        \includegraphics[width=\linewidth]{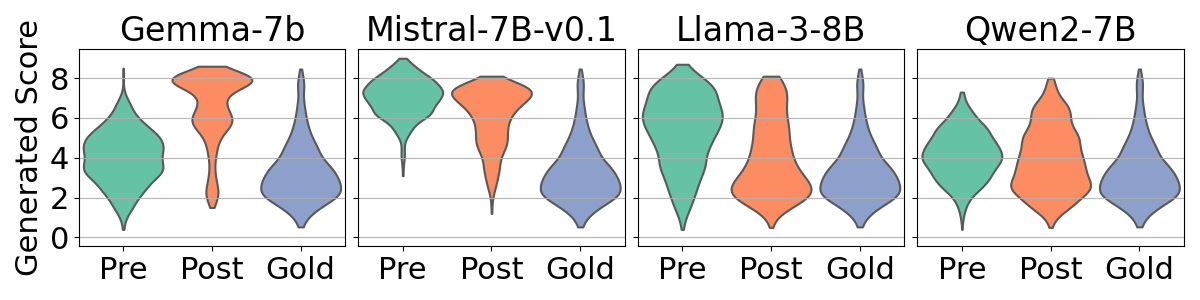}
        \subcaption{Nepali to English (Ne-En)}
    \end{minipage}
    \begin{minipage}[b]{\linewidth}
        \centering
        \includegraphics[width=\linewidth]{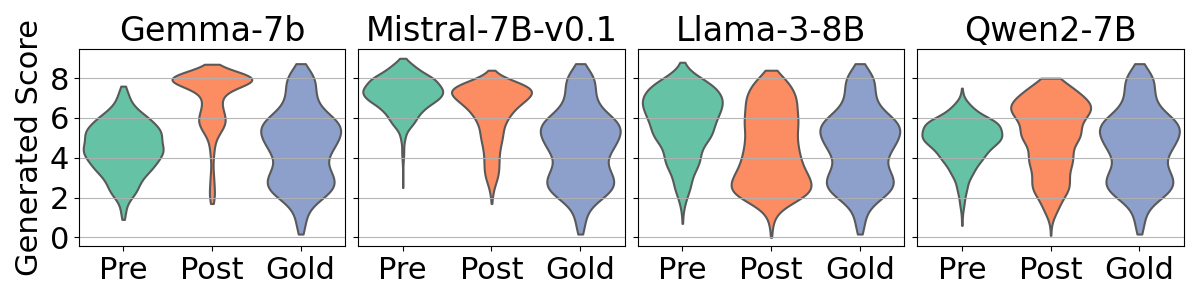}
        \subcaption{Sinhala to English (Si-En)}
    \end{minipage}
    \caption{MTQE score distribution.}
    \label{fig:mtqe_0-9}
\end{figure}

Evaluators are primarily post-alignment LLMs, which undergo alignment methods such as instruction tuning~\cite{wei2022finetuned} and preference tuning~\cite{NEURIPS2022-b1efde53, rafailov2023direct}.
\emph{Alignment} adjusts models to generate responses aligned with human intentions and values, improving instruction-following and evaluation performance.
However, existing studies have pointed out concerns about potential ``side effects'' of alignment.
Specifically, alignment can bias model outputs toward certain patterns, potentially reducing diversity \cite{pmlr-v202-santurkar23a} and creativity \cite{mohammadi2024creativityleftchatprice}.

Building on this background, this study hypothesizes that numerical bias in LLM evaluators is also a side effect of alignment and investigates the impact of LLM alignment on numerical bias by comparing the evaluation score distributions and evaluation performance of pre- and post-alignment models (\S\ref{sec:rq1}).
Our findings indicate that alignment amplifies numerical bias and harms evaluation performance (as shown in Figure~\ref{fig:mtqe_0-9}).
However, despite this bias, the post-alignment model still outperforms the pre-alignment model.
This suggests that while alignment introduces numerical bias, the resulting performance degradation is generally not severe enough to outweigh the benefits of improved instruction-following ability.
Therefore, there is still room improve post-alignment model performance by mitigating the bias.

To address this challenge, we investigate potential solutions to mitigate numerical bias and evaluate the effectiveness of three approaches: modifying the temperature, calibrating the output distribution, and adjusting the score range (\S\ref{sec:rq2}).
% Our experiments show that adjusting the score range is the most effective, as it not only reduces numerical bias but also improves evaluation performance.
Our experiments show that adjusting the score range can reduce numerical bias and, in some cases, enhance correlation with human judgments.
These findings suggest that the score range, which has often been treated as a fixed design choice in prior work, should instead be reconsidered as a tunable hyperparameter in LLM-as-a-judge settings.

Our key findings are as follows:
\begin{itemize}
    \setlength{\itemsep}{-3pt}
    \item Numerical bias appears in LLM outputs after alignment, though it was weaker or absent before. The bias is stronger for high-resource languages than for low-resource languages.
    \item The strength of numerical bias varies by model, and stronger bias correlates with lower evaluation accuracy. This suggests that users can select LLM evaluators based on bias strength, even without gold data.
    \item Adjusting the score range in the prompt mitigates numerical bias and improves evaluation accuracy. Optimizing the score range is key to maximizing accuracy with the same model.
\end{itemize}

\section{Related Work}
Several studies have demonstrated the effectiveness of LLMs as evaluators. 
\citet{chiang-lee-2023-large} conducted experiments on a story and adversarial sample evaluation, showing that LLM assessments align with human experts.
Their study highlights advantages like high reproducibility and low cost but notes limitations in fact-dependent evaluations and potentially harmful responses.
The effectiveness of LLMs as evaluators has been demonstrated in the tasks examined in this study.
\citet{kocmi-federmann-2023-gemba, kocmi-federmann-2023-large} demonstrated that using GPT-3.5 and GPT-4 as evaluators in MTQE outperformed existing methods.
\citet{kobayashi-etal-2024-large} found similar results when using GPT-4 as an evaluator for GECQE.
Furthermore, \citet{enomoto-etal-2024-tmu} showed that employing GPT-4 as an evaluator achieved first place in an LCP shared task, reinforcing its effectiveness in evaluation.
These findings support the viability of LLMs as evaluators for diverse linguistic tasks.

Despite their effectiveness, LLM evaluators exhibit biases that affect reliability.
\citet{stureborg2024largelanguagemodelsinconsistent} identified three key biases in the LLM-as-a-judge: (i) a preference for lower-perplexity texts, (ii) improved evaluation performance with a smaller output value range, and (iii) influence from prior evaluations in multi-instance assessments.
Their study quantified these biases and proposed mitigation guidelines.
\citet{ohi-etal-2024-likelihood} observed a likelihood bias in scoring-based evaluations for data-to-text and grammatical error correction tasks, where LLMs favor texts with higher likelihoods.
They showed this bias can be reduced using biased instances as few-shot examples.
While these studies provide insights into LLM evaluation biases, they do not address numerical bias, where certain values are overused unnaturally.
Our study focuses on this issue, analyzing its causes and hypothesizing its link to alignment.

\section{Analysis of Alignment Influence}
\label{sec:rq1}
As discussed in the Introduction, this study aims to uncover the causes of numerical bias.
Specifically, we hypothesize that alignment processes induce numerical bias because previous studies suggest that alignment reduces the diversity of model outputs\cite{pmlr-v202-santurkar23a, mohammadi2024creativityleftchatprice}.
In line with this, our first research question—\textbf{RQ1: How does LLM alignment influence numerical bias?}—guides our investigation.
Based on this hypothesis, this section examines how alignment influences numerical bias.

\subsection{Experimental Design for RQ1} 
To verify whether alignment induces numerical bias, we compare the distributions of output scores from pre- and post-alignment models.
Specifically, we visualize these distributions and measure the kurtosis of the output scores, as kurtosis serves as a metric for numerical bias\footnote{Kurtosis is more appropriate than variance or interquartile range for quantifying local concentration around specific values rather than overall distribution narrowness.}.
If the kurtosis of the post-alignment model is higher than that of the pre-alignment model, it indicates that alignment increases numerical bias.

Additionally, we evaluate the performance of both pre- and post-alignment models on scoring (regression) tasks, such as MTQE, by measuring Pearson's correlation coefficient $r$ between human scores and the model's scores.
Based on the results of the distribution and evaluation performance comparisons, we analyze how alignment affects numerical bias and evaluation performance.

\subsection{Experimental Setup for RQ1} 
\subsubsection{Data}
\label{sec:tasks}
The target evaluation tasks are the following three regression tasks: MTQE, GECQE, and LCP.
In these tasks, LLM takes text as input and outputs the evaluation results as a numerical score.

\paragraph{MTQE}
MTQE aims to automatically evaluate the quality of system-generated translations without relying on reference translations.
The resulting quality scores indicate the reliability of a translation system and the potential need for human post-editing.
Quality estimation can be performed at different granularity, including word, phrase, sentence, and document levels.
In this study, we use sentence-level\footnote{It can evaluate translations with contextual awareness while also being efficient for LLM-based evaluation in terms of processing speed.} Direct Assessment (DA) \cite{fomicheva-etal-2022-mlqe} scores from the WMT 2020 Shared Task on Quality Estimation \cite{specia-etal-2020-findings-wmt} for seven language pairs\footnote{En-De, En-Zh, Et-En, Ne-En, Ro-En, Ru-En, Si-En}.
We instruct the LLM to output scores within the specified range and analyze numerical bias.
The distribution of gold scores varies across language pairs, reflecting the unique challenges and quality variability of translations in each language.
This diversity serves as a critical benchmark for evaluating whether the model can accurately capture differences in translation quality.

\paragraph{GECQE}
GECQE aims to automatically evaluate the quality of GEC system outputs without reference.
The goal is to assess the quality of the corrected sentence is and whether further revisions are necessary.
We use TMUGFM \cite{yoshimura-etal-2020-reference} and autoJQE \cite{suzuki-etal-2022-construction}.
TMUGFM focuses on English GEC and provides human evaluations for grammaticality, fluency, and meaning preservation.
We use grammaticality, which plays the most critical role in overall evaluation, as the overall evaluation score.
Conversely, autoJQE provides manually assigned quality scores for two Japanese GEC datasets: FLUTEC\footnote{\url{https://github.com/kiyama-hajime/FLUTEC}} and TECJL \cite{koyama-etal-2020-construction}.

\paragraph{LCP}
LCP automatically predicts the complexity of words in a sentence, given a sentence and a target word pair.
Accurate lexical complexity prediction helps guide users to suitable texts or adapt content to their needs.
There are two complexity estimation types: single-word and multi-word, with this study focusing on single-word complexity.
For this study, we use test data from the MLSP2024 shared task \cite{shardlow-etal-2020-complex}, covering nine languages\footnote{Ca, En, Fil, Fr, De, It, Si, Es, Ja}.
All LCP results are presented in Appendix \ref{sec:appendix-result}.

\subsubsection{Score Generation Procedure}
\label{sec:score-generation}
{\setlength{\tabcolsep}{4.5pt}
\begin{table}[t]
    \centering
    \small
    \begin{tabular}{lll}
    \toprule
        Series & Pre-alignment & Post-alignment \\
    \midrule
        Gemma & gemma-7b & gemma-7b-it \\
        Mistral & Mistral-7B-v0.1 & Mistral-7B-Instruct-v0.1 \\
        Llama 3  & Meta-Llama-3-8B & Meta-Llama-3-8B-Instruct \\
        Qwen 2 & Qwen2-7B & Qwen2-7B-Instruct \\
    \bottomrule
    \end{tabular}
    \caption{List of LLM evaluators used in the experiments.}
    \label{tab:llm-list}
\end{table}}

We explain the procedure for text evaluation with LLM-generated scores, taking the MTQE task as an example.
In the experiments, we use four open LLMs listed in Table \ref{tab:llm-list}.
To suppress non-numeric outputs, we set \textit{max\_token=5}.
For the experiments addressing RQ1, the temperature is fixed at $0.7$.

\paragraph{Step 1: Prompt Design}  
\begin{table}[t]
    \centering
    \small
    \begin{tabular}{p{7.25cm}}
        \toprule
        Please analyze the given source and translated sentences and output a translation quality score on a continuous scale ranging from \{\{min score\}\} to \{\{max score\}\}.\\
        Translation quality should be evaluated based on both fluency and adequacy.\\
        A score close to \{\{min score\}\} indicates a low quality translation, while a score close to
        \{\{max score\}\} indicates a high quality translation.\\
        Do not provide any explanations or text apart from the score.\\
        
        \{\{source language\}\} 
        Sentence: \{\{$src_i$\}\} \\
        \{\{target language\}\} Sentence: \{\{$hyp_i$\}\} \\
        Score:\\
        \bottomrule
    \end{tabular}
    \caption{Prompt Template for MTQE.}
    \label{tab:prompt-template}
\end{table}

We use the prompt template shown in Table \ref{tab:prompt-template}\footnote{Prompts for other tasks are shown in the Appendix \ref{sec:appendix-prompt}.}.
Based on GEMBA \cite{kocmi-federmann-2023-large}, the following arguments are configured for each dataset and embedded into the prompt template:
\begin{itemize}
    \setlength{\itemsep}{-3pt}
    \item source language name: \{\{source language\}\}
    \item target language name: \{\{target language\}\}
    \item min value of the score range: \{\{min score\}\}  
    \item max value of the score range: \{\{max score\}\}  
    \item source sentences: $src_{1..N}$
    \item translated sentences: $hyp_{1..N}$  
\end{itemize}
For RQ1, the score range is $0\text{--}9$ for all tasks\footnote{To match the evaluation range specified for the model, gold scores are rescaled before calculating kurtosis.}.

\paragraph{Step 2: Calculation of Evaluation Scores}
Using the prompt from Step 1, numerical values are generated 10 times.
From the 10 generated results, non-numeric tokens are excluded.
For numeric values outside the specified range, clipping is applied to bring them within the range by replacing them with the upper or lower boundary value.
The final score is the average of the processed scores.

\subsection{Experimental Results for RQ1}
{\setlength{\tabcolsep}{5.0pt}
\begin{table}[t]
    \centering
    \small
    \renewcommand{\arraystretch}{0.9}
    \begin{tabular}{llrrrrr}
    \toprule
        & & \multicolumn{3}{c}{Kurtosis} & \multicolumn{2}{c}{$r$} \\
        \cmidrule(lr){3-5}
        \cmidrule(lr){6-7}
        \multicolumn{1}{c}{Lang} & \multicolumn{1}{c}{Model} & \multicolumn{1}{c}{gold} & \multicolumn{1}{c}{pre} & \multicolumn{1}{c}{post} & \multicolumn{1}{c}{pre} & \multicolumn{1}{c}{post} \\
    \midrule
        \multirow{4}{*}{En-De} & Gemma & 1.48 & \textbf{0.27} & 128.17 & \textbf{0.33$^{*}$} & 0.08\hphantom{*} \\
        & Mistral & 1.48 & \textbf{0.41} & 52.92 & 0.11$^{*}$ & \textbf{0.20$^{*}$} \\
        & Llama & 1.48 & \textbf{2.53} & 21.05 & 0.33$^{*}$ & \textbf{0.35$^{*}$} \\
        & Qwen & 1.48 & \textbf{0.53} & 21.05 & 0.33$^{*}$ & \textbf{0.38$^{*}$} \\
    \midrule
        \multirow{4}{*}{En-Zh} & Gemma & 0.20 & \textbf{\textminus 0.06} & 21.57 & \textbf{0.30$^{*}$} & 0.23$^{*}$ \\
        & Mistral & 0.20 & \textbf{0.73} & 11.33 & 0.04\hphantom{*} & \textbf{0.22$^{*}$} \\
        & Llama & 0.20 & \textbf{\textminus 0.06} & 3.96 & 0.27$^{*}$ & \textbf{0.37$^{*}$} \\
        & Qwen & 0.20 & \textbf{\textminus 0.05} & \textminus 0.84 & 0.34$^{*}$ & \textbf{0.41$^{*}$} \\
    \midrule
        \multirow{4}{*}{Et-En} & Gemma & \textminus 1.24 & \textbf{\textminus 0.11} & 11.64 & \textbf{0.45$^{*}$} & 0.37$^{*}$ \\
        & Mistral & \textminus 1.24 & \textbf{0.53} & 4.93 & 0.23$^{*}$ & \textbf{0.49$^{*}$} \\
        & Llama & \textminus 1.24 & \textbf{0.98} & \textminus 1.09 & 0.55$^{*}$ & \textbf{0.66$^{*}$} \\
        & Qwen & \textminus 1.24 & 0.47 & \textbf{\textminus 0.32} & 0.55$^{*}$ & \textbf{0.61$^{*}$} \\
    \midrule
        \multirow{4}{*}{Ne-En} & Gemma & 0.97 & \textbf{\textminus 0.34} & 0.45 & 0.21$^{*}$ & \textbf{0.37$^{*}$} \\
        & Mistral & 0.97 & 0.58 & \textbf{\textminus 0.19} & 0.17$^{*}$ & \textbf{0.40$^{*}$} \\
        & Llama & 0.97 & \textbf{\textminus 0.64} & \textminus 0.95 & 0.32$^{*}$ & \textbf{0.45$^{*}$} \\
        & Qwen & 0.97 & \textbf{\textminus 0.39} & \textminus 0.77 & 0.44$^{*}$ & \textbf{0.52$^{*}$} \\
    \midrule
        \multirow{4}{*}{Ro-En}& Gemma & \textminus 0.16 & \textbf{\textminus 0.59} & 2.73 & 0.49$^{*}$ & \textbf{0.59$^{*}$} \\
        & Mistral & \textminus 0.16 & \textbf{0.71} & 1.12 & 0.30$^{*}$ & \textbf{0.66$^{*}$} \\
        & Llama & \textminus 0.16 & \textbf{0.56} & \textminus 1.45 & 0.57$^{*}$ & \textbf{0.77$^{*}$} \\
        & Qwen & \textminus 0.16 & \textbf{0.10} & \textminus 0.84 & 0.74$^{*}$ & \textbf{0.78$^{*}$} \\
    \midrule
        \multirow{4}{*}{Ru-En} & Gemma & \textminus 0.57 & \textbf{0.40} & 9.70 & \textbf{0.31$^{*}$} & 0.23$^{*}$ \\
        & Mistral & \textminus 0.57 & \textbf{0.17} & 6.33 & 0.19$^{*}$ & \textbf{0.37$^{*}$} \\
        & Llama & \textminus 0.57 & 3.95 & \textbf{3.11} & 0.13$^{*}$ & \textbf{0.45$^{*}$} \\
        & Qwen & \textminus 0.57 & 1.81 & \textbf{1.80} & 0.46$^{*}$ & \textbf{0.53$^{*}$} \\
    \midrule
        \multirow{4}{*}{Si-En} & Gemma & \textminus 0.72 & \textbf{\textminus 0.37} & 1.97 & 0.24$^{*}$ & \textbf{0.38$^{*}$} \\
        & Mistral & \textminus 0.72 & 1.38 & \textbf{0.03} & 0.12$^{*}$ & \textbf{0.35$^{*}$} \\
        & Llama & \textminus 0.72 & \textbf{\textminus 0.36} & \textminus 1.13 & 0.36$^{*}$ & \textbf{0.47$^{*}$} \\
        & Qwen & \textminus 0.72 & \textbf{0.48} & \textminus 0.85 & 0.45$^{*}$ & \textbf{0.48$^{*}$} \\
    \bottomrule
    \end{tabular}
    \caption{Kurtosis of MTQE score distributions and Pearson correlation coefficient ($r$). For each language pair and model, the \underline{smallest absolute kurtosis} and the \underline{highest $r$} are highlighted in \textbf{bold}. $^{*}$ indicate statistically significant correlations ($p<0.01$).}
    \label{tab:mtqe_0-9}
\end{table}}

\begin{figure}[t]
    \centering
    \begin{minipage}[b]{\linewidth}
        \centering
        \includegraphics[width=\linewidth]{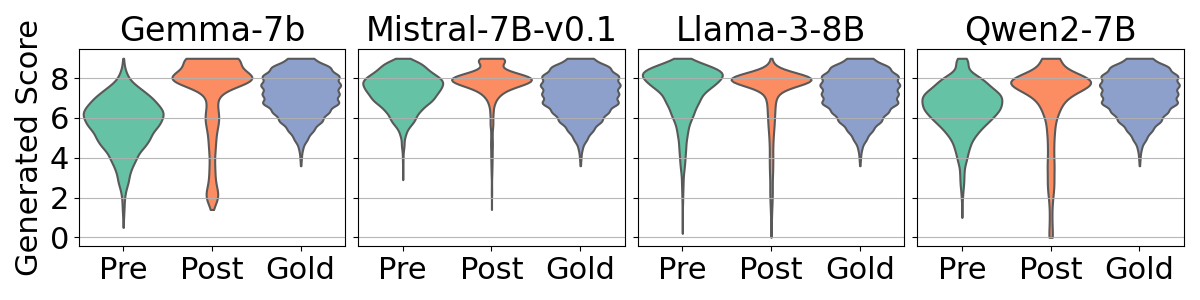}
        \subcaption{TMUGFM (English)}
    \end{minipage}
        \begin{minipage}[b]{\linewidth}
        \centering
        \includegraphics[width=\linewidth]{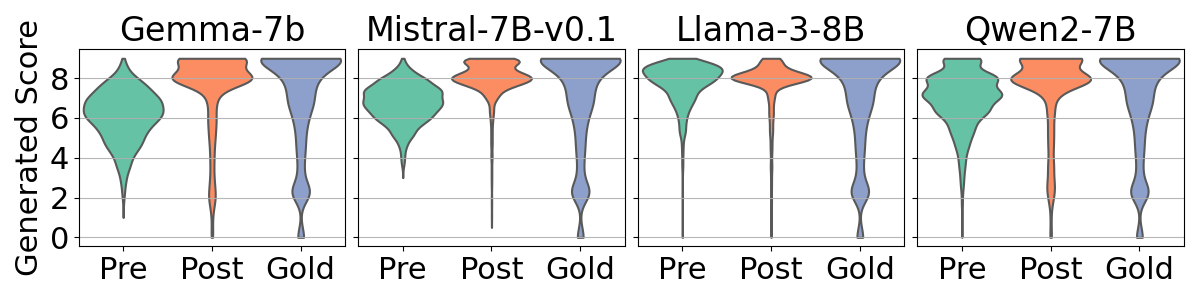}
        \subcaption{TECJL (Japanese)}
    \end{minipage}
    \caption{Distribution of GECQE scores.}
    \label{fig:gecqe_0-9}
\end{figure}

{\setlength{\tabcolsep}{4.0pt}
{\begin{table}[t]
    \centering
    \small
    \renewcommand{\arraystretch}{0.9}
    \begin{tabular}{llrrrrr}
    \toprule
        & & \multicolumn{3}{c}{Kurtosis} & \multicolumn{2}{c}{$r$ } \\
        \cmidrule(lr){3-5}
        \cmidrule(lr){6-7}
        \multicolumn{1}{c}{Data} & \multicolumn{1}{c}{Model} & \multicolumn{1}{c}{gold} & \multicolumn{1}{c}{pre} & \multicolumn{1}{c}{post} & \multicolumn{1}{c}{pre} & \multicolumn{1}{c}{post} \\
    \midrule
        \multirow{4}{*}{TMUGFM} & Gemma & \textminus0.47 & \textbf{0.32} & 1.03 & 0.22$^{*}$ & \textbf{0.43$^{*}$} \\
        & Mistral & \textminus0.47 & \textbf{0.34} & 8.49 & 0.03\hphantom{*} & \textbf{0.25$^{*}$} \\
        & Llama & \textminus0.47 & \textbf{2.23} & 3.51 & \textbf{0.32$^{*}$} & 0.18$^{*}$ \\
        & Qwen & \textminus0.47 & \textbf{0.91} & 2.20 & 0.31$^{*}$ & \textbf{0.37$^{*}$} \\
    \midrule
        \multirow{4}{*}{FLUTEC} & Gemma & \textminus0.55 & \textbf{0.15} & 4.41 & 0.25$^{*}$ & \textbf{0.49$^{*}$} \\
        & Mistral & \textminus0.55 & \textbf{0.14} & 14.13 & \textminus0.01\hphantom{*} & \textbf{0.29$^{*}$} \\
        & Llama & \textminus0.55 & \textbf{3.15} & 11.72 & \textbf{0.14$^{*}$} & 0.12$^{*}$ \\
        & Qwen & \textminus0.55 & \textbf{0.59} & 3.27 & \textbf{0.53$^{*}$} & 0.52$^{*}$ \\
    \midrule
        \multirow{4}{*}{TECJL} & Gemma & 0.40 & \textbf{0.09} & 3.04 & 0.28$^{*}$ & \textbf{0.44$^{*}$} \\
        & Mistral & 0.40 & \textbf{0.08} & 12.73 & 0.03\hphantom{*} & \textbf{0.32$^{*}$} \\
        & Llama & 0.40 & \textbf{8.01} & 9.10 & \textbf{0.30$^{*}$} & 0.21$^{*}$ \\
        & Qwen & 0.40 & \textbf{1.11} & 2.54 & 0.56$^{*}$ & \textbf{0.57$^{*}$} \\
    \bottomrule
    \end{tabular}
    \caption{Kurtosis of GECQE score distributions and Pearson correlation coefficient ($r$). $^{*}$ indicate statistically significant correlations ($p<0.01$).}
    \label{tab:gecqe_0-9}
\end{table}}

\begin{figure}[t]
    \centering
    \includegraphics[width=\linewidth]{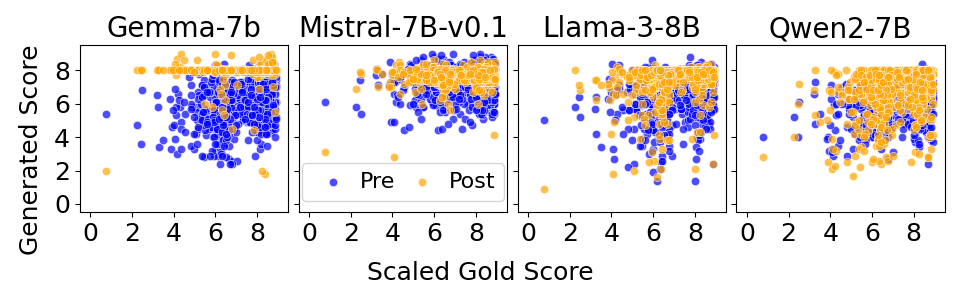}
    \caption{Scatter plots of MTQE score for En-De.}
    \label{fig:scatter_mtqe_0-9}
\end{figure}

\paragraph{Alignment induces numerical bias.}
Figures \ref{fig:mtqe_0-9} and \ref{fig:gecqe_0-9} illustrate the distribution of evaluation scores within the score range of $0\text{--}9$ in the MTQE and GECQE tasks, respectively\footnote{Results for MTQE (other languages) and GECQE (FLUTEC) are shown in Appendix Figures \ref{fig:mtqe_0-9_appendix} and \ref{fig:gecqe_0-9_appendix}.}.
These figures show the same trends.
Specifically, in pre-alignment models, score distributions are more varied, reflecting diversity across models and language pairs.
This result—the apparent absence of bias in the pre-alignment model—may be attributed to its weak instruction-following ability, leading to random output scores.
Conversely, post-alignment models generate scores that are heavily concentrated around $8$, suggesting that alignment introduces excessive consistency.
Notably, while the gold score distributions vary across data, post-alignment models produce evaluation score distributions with strikingly similar shapes.
This suggests that post-alignment models may fail to adequately capture differences in translation quality, leading to numerical bias, where scores cluster excessively around specific values, regardless of the input.

Tables \ref{tab:mtqe_0-9} and \ref{tab:gecqe_0-9} present kurtosis of evaluation score distributions.
Higher kurtosis indicates a sharper distribution and stronger numerical bias.
Across most datasets, the post-alignment models exhibit higher kurtosis compared to their pre-alignment counterparts, further supporting the presence of numerical bias induced by alignment.

\paragraph{Models with stronger numerical bias tend to have lower evaluation accuracy.}
Tables \ref{tab:mtqe_0-9} and \ref{tab:gecqe_0-9} further confirm that models and language pairs with higher kurtosis tend to have lower evaluation accuracy, as seen in their Pearson correlation coefficient ($r$) with gold scores.
Kurtosis and $r$ show a strong negative correlation of $-0.60$ for both MTQE and GECQE, indicating that greater numerical bias leads to lower accuracy.
Although alignment improves instruction-following abilities and can enhance evaluation accuracy, excessive numerical bias can undermine its benefits.

To investigate the impact of numerical bias on evaluation accuracy, we created scatter plots of LLM-generated scores (vertical axis) against human evaluation scores (horizontal axis), as shown in Figure \ref{fig:scatter_mtqe_0-9}\footnote{Results for GECQE and LCP are in Appendix Figure \ref{fig:scatter_0-9}.}.
Pre-alignment models exhibit low kurtosis and can be considered unbiased. 
However, the points in the scatter plot do not align with the $y = x$ line, indicating that their evaluation task performance is not necessarily high.

\paragraph{The strength of bias varies across models.}
As shown in Figures \ref{fig:mtqe_0-9}, along with Table \ref{tab:mtqe_0-9}, bias varies considerably across models.
Post-alignment Gemma exhibits the highest kurtosis for most language pairs, indicating the strongest numerical bias.
Mistral follows Gemma, with generally high kurtosis and particularly strong bias for En-De and En-Zh.
In contrast, Llama and Qwen show lower kurtosis for most language pairs except En-De compared to the other two models.

\paragraph{The strength of bias varies across languages.}
The degree of numerical bias introduced by alignment varies across language pairs.
For En-De, En-Zh, Et-En, Ro-En, and Ru-En, post-alignment models exhibit highly skewed distributions, with scores heavily concentrated on specific values.
In contrast, for Ne-En and Si-En, distributions differ among post-alignment models, with no excessive concentration on specific values. 
Compared to high-resource languages like English (En), German (De), and Chinese (Zh), which have stronger biases, Nepali (Ne) and Sinhala (Si) are low-resource languages.
Their lower representation in alignment data may contribute to differences in how alignment affects numerical bias.

\section{Assessing Solutions for Bias Mitigation}
\label{sec:rq2}

The experiments addressing RQ1 confirmed that alignment introduces numerical bias in the evaluation score distribution.
However, despite the absence of numerical bias, pre-alignment models demonstrate lower evaluation task performance compared to their post-alignment counterparts.
Thus, post-alignment models remain effective as evaluator LLMs, even in the presence of numerical bias.
Mitigating numerical bias in post-alignment models has the potential to further enhance evaluation task performance.
Based on \textbf{RQ2: What is a potential solution to mitigate numerical bias?}, we explore solutions that might further improve evaluation task performance by reducing the numerical bias in the post-alignment model.

\subsection{Experimental Design for RQ2}
To mitigate bias and boost evaluation performance, we investigate two approaches: adjustment of the output distribution—using techniques such as temperature scaling and distribution calibration—and adjustment of the score range.

\begin{figure}[t]
    \small
    \centering
    \setlength\abovecaptionskip{2truemm}
    
    \includegraphics[width=1\linewidth]{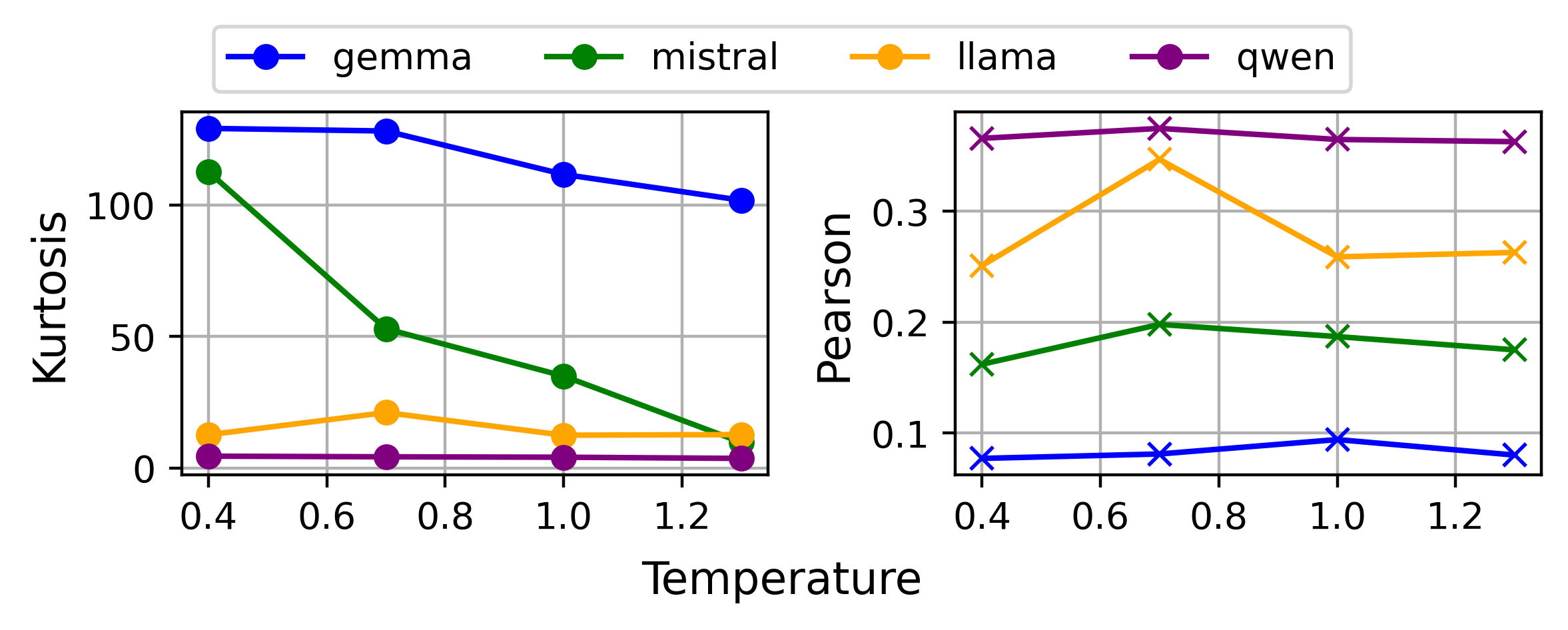}
    \caption{Kurtosis and Pearson correlation coefficient ($r$) for each temperature in MTQE (En-De).}
    \label{fig:mtqe-temperature}
\end{figure}

\begin{figure}[t]
    \small
    \centering
    \setlength\abovecaptionskip{2truemm}
    
    \includegraphics[width=1\linewidth]{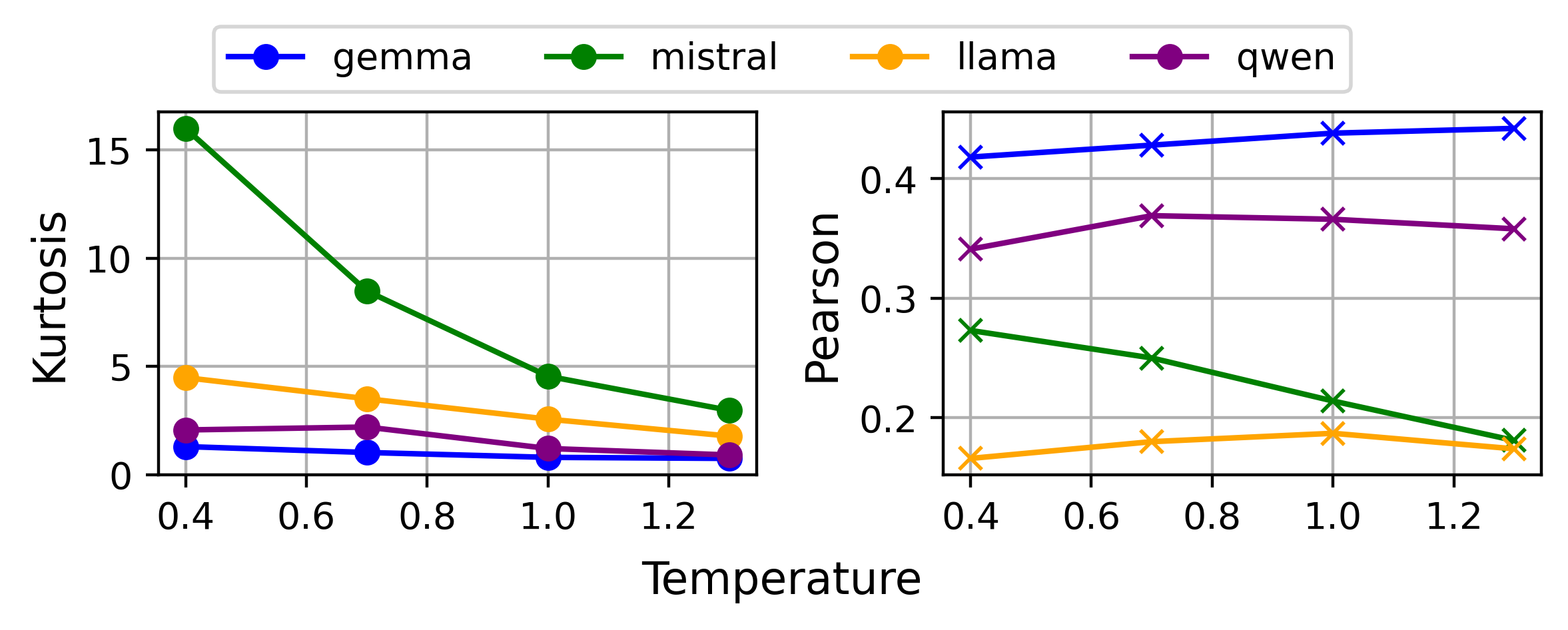}
    \caption{Kurtosis and Pearson correlation coefficient ($r$) for each temperature in GECQE (TMUGFM).}
    \label{fig:gecqe-temperature}
\end{figure}

\begin{table}[t]
    \centering
    \small
    \setlength\abovecaptionskip{2truemm}
    
    \renewcommand{\arraystretch}{0.9}
    \begin{tabular}{lrrrr}
        \toprule
        & \multicolumn{2}{c}{Kurtosis} & \multicolumn{2}{c}{$r$} \\
         \cmidrule(lr){2-3} \cmidrule(lr){4-5}
        \textbf{} & original & calibrated & original & calibrated \\
        \midrule
        Gemma   & 128.174  & \textbf{47.932}  & \textbf{0.081}  & 0.076  \\
        Mistral & 52.923   & \textbf{3.394}  & \textbf{0.198}  & 0.147  \\
        Llama   & 21.051   & \textbf{11.886}  & 0.347  & \textbf{0.380}  \\
        Qwen    & 4.213    & \textbf{2.704}   & 0.375  & \textbf{0.377}  \\
        \bottomrule
    \end{tabular}
    \caption{Kurtosis and Pearson correlation coefficients ($r$) before and after calibration of the distribution in MTQE (En-De).}
    \label{tab:calibration_mtqe}
\end{table}

\begin{table}[t]
    \centering
    \small
    \setlength\abovecaptionskip{2truemm}
    
    \renewcommand{\arraystretch}{0.9}
    \begin{tabular}{lrrrr}
        \toprule
        & \multicolumn{2}{c}{Kurtosis} & \multicolumn{2}{c}{$r$} \\
         \cmidrule(lr){2-3} \cmidrule(lr){4-5}
        \textbf{} & original & calibrated & original & calibrated \\
        \midrule
        Gemma   & \textbf{1.028}  & 2.092  & 0.428  & \textbf{0.430}  \\
        Mistral & 8.485 & \textbf{0.106} & 0.250 & \textbf{0.297}  \\
        Llama   & 3.510 & \textbf{0.994} & \textbf{0.180} & 0.021 \\
        Qwen    & \textbf{2.204} & 8.156 & \textbf{0.369} & 0.353  \\
        \bottomrule
    \end{tabular}
    \caption{Kurtosis and Pearson correlation coefficients ($r$) before and after calibration of the distribution in GECQE (TMUGFM).}
    \label{tab:calibration_gecqe}
\end{table}

\paragraph{Adjustment of output distribution}
The most straightforward approach to reducing bias is to adjust the output distribution.
Appropriately calibrating the distribution yields a more diverse range of numerical outputs and mitigates numerical bias.
Therefore, this study explores bias mitigation by adjusting the output distribution.
Specifically, as outlined below, we attempt to modify the temperature parameter and apply a calibration method to adjust the distribution.

The simplest way to adjust the distribution to obtain diverse outputs is to adjust the temperature parameter.
The temperature parameter is a hyperparameter used in probabilistic text generation models, to control the diversity of the output.
It scales the logits before applying softmax function, influencing the probability distribution of the next token selection.
Increasing the temperature parameter is expected to produce more diverse numerical output as the output distribution approaches uniformity.

Another way to mitigate numerical bias is to adopt the calibration method proposed by \citet{jiang-etal-2023-generative}.
Their generative calibration approach adjusts the in-context predictive distribution by recalibrating the label marginal \( p(y) \), estimated via Monte-Carlo sampling over LLM outputs.
Since numerical bias can be viewed as a label shift induced by the alignment data distribution, we apply this approach to correct the distribution of numeric tokens in LLM-based evaluation.
This method calibrates the probability distribution of score $y$ given input $x$, denoted as $p(y \mid x)$, using the language model distribution $p(y)$ and the true data distribution $q(y)$.
In this study, to approximate $p(y \mid x)$, we sample $n$ evaluation scores per input\footnote{$p(y \mid x)$ can be obtained via softmax or logits, but approximating it through sampling allows application to closed models.
We use $n$ valid numerical scores from the evaluation scores produced by the method described in \S\ref{sec:score-generation}.} and compute the final score $\hat{y}$ as a weighted average:
\begin{equation}
\hat{y} = \sum_{i=1}^{n} w_i y_i, \quad w_i = \frac{q(y_i)}{p(y_i)} \Bigg/ \sum_{j=1}^{n} \frac{q(y_j)}{p(y_j)}
\end{equation}
where $p(y)$ is estimated by sampling $1,000$ times, and $q(y)$ is modeled as a Beta distribution\footnote{Due to the asymmetry of gold scores, the Beta distribution is more appropriate than the normal distribution for this tasks.} with parameters $\alpha$ and $\beta$ estimated via maximum likelihood.
The marginal probability $q(y)$ is computed by integrating the Beta density function over the appropriate range\footnote{Since $y$ is an integer, the probability density is integrated over $(y, y+1]$ for $y>0$ and $[0,1]$ for $y=0$.}.
This sampling-based approach reduces computational cost while effectively calibrating score distributions.

\paragraph{Adjustment of output range}
Modifying the evaluation score range specified in the prompt alters the set of possible output tokens.
\citet{stureborg2024largelanguagemodelsinconsistent} found that expanding the score range reduced evaluation accuracy.
They cautioned against the possibility of numerical bias from expanding the score range.
However, they did not empirically validate whether a direct relationship exists between the score range and numerical bias.
Thus, we test the hypothesis that score range adjustments influence evaluation accuracy by including or excluding biased scores.
By quantifying numerical bias using the kurtosis of the distribution, we can investigate the impact of score range adjustments on numerical bias and evaluation performance.

\subsection{Experimental Setup for RQ2}
We analyze En-De for MTQE and TMUGFM for GECQE as data for which bias is significant.
When testing the effect of temperature scaling, the score range is fixed at $0\text{--}9$ and the temperature values are varied by four values $\{0.4, 0.7, 1.0, 1.3\}$.
When testing the effect of score range adjustment, fix the temperature value at $0.7$ and specify a score range of $\{1\text{--}5, 0\text{--}9, 1\text{--}100\}$.

\subsection{Experimental Results for RQ2}
\paragraph{Temperature Scaling and Distribution Calibration are effective in mitigating numerical bias.}

Results of temperature scaling are shown in Figures~\ref{fig:mtqe-temperature} and \ref{fig:gecqe-temperature}.
This suggests that setting a higher temperature parameter can serve as a mitigation strategy to reduce the impact of alignment-induced numerical bias.
The strategy is particularly effective for models with stronger bias.
However, excessively high temperatures may decrease evaluation accuracy.
In this experiment, the optimal temperature for maximizing $r$ is 1.0 for Gemma and 0.7 for the other models, aligning with their default settings.
Thus, when prioritizing evaluation accuracy, using the default temperature values is most effective.
Temperature scaling mitigated bias but did not necessarily improve evaluation performance.

Tables~\ref{tab:calibration_mtqe} and \ref{tab:calibration_gecqe} show the effects of calibration on kurtosis and Pearson correlation in MTQE and GECQE.
In MTQE, all models exhibit reduced kurtosis, and in GECQE, Mistral and Llama—both of which initially had relatively high kurtosis—also show improvement.
This indicates that distribution calibration is effective in mitigating numerical bias.
However, its impact on evaluation performance varies.
In the models whose original accuracy is relatively high, such as Llama and Qwen, calibration results in a slight improvement.
Conversely, when the original accuracy is low, calibration tends to further degrade performance.
This suggests that while calibration reshapes the score distribution, it does not always improve performance.

\begin{table}[t]
\small
\setlength\abovecaptionskip{2truemm}

\renewcommand{\arraystretch}{0.9}
\centering
\begin{tabular}{lrrrrrr}
    \toprule
    & \multicolumn{3}{c}{Kurtosis} & \multicolumn{3}{c}{ $r$ } \\
    \cmidrule(lr){2-4} \cmidrule(lr){5-7}
    Model & \multicolumn{1}{c}{1--5} & \multicolumn{1}{c}{0--9} & \multicolumn{1}{c}{1--100} & \multicolumn{1}{c}{1--5} & \multicolumn{1}{c}{0--9} & \multicolumn{1}{c}{1--100} \\
    \midrule
    Gemma & \textbf{71.79} & 128.17 & 87.45 & 0.05 & 0.08 & \textbf{0.14} \\
    Mistral & \textbf{11.95} & 52.92 & 42.00 & \textbf{0.22} & 0.20 & 0.21 \\
    Llama & \textbf{10.19} & 21.05 & 23.96 & 0.27 & \textbf{0.35} & 0.33 \\
    Qwen & 6.87 & \textbf{4.21} & 5.28 & 0.34 & \textbf{0.38} & 0.37 \\
    \bottomrule
    \end{tabular}
\caption{Kurtosis and Pearson correlation coefficient ($r$) for each score ranges in MTQE (En-De) .}
\label{tab:mtqe-score-scale}
\end{table}

\begin{table}[t]
\small
\setlength\abovecaptionskip{2truemm}
\renewcommand{\arraystretch}{0.9}
\centering
\begin{tabular}{lrrrrrr}
    \toprule
    & \multicolumn{3}{c}{Kurtosis} & \multicolumn{3}{c}{$r$} \\
    \cmidrule(lr){2-4} \cmidrule(lr){5-7}
    Model & \multicolumn{1}{c}{1--5} & \multicolumn{1}{c}{0--9} & \multicolumn{1}{c}{1--100} & \multicolumn{1}{c}{1--5} & \multicolumn{1}{c}{0--9} & \multicolumn{1}{c}{1--100} \\
    \midrule
    Gemma & \textbf{\textminus0.58} & 1.03 & 5.38 & \textbf{0.45} & 0.43 & 0.44 \\
    Mistral & \textbf{1.95} & 8.49 & 12.09 & 0.33 & 0.25 & \textbf{0.33} \\
    Llama & \textbf{0.73} & 3.51 & 2.24 & \textbf{0.25} & 0.18 & 0.18 \\
     Qwen & \textbf{2.13} & 2.20 & 4.63 & 0.35 & \textbf{0.37} & 0.30 \\
    \bottomrule
\end{tabular}
\caption{Kurtosis and Pearson correlation coefficient ($r$) for each score ranges in GECQE (TMUGFM) .}
\label{tab:gecqe-score-scale}
\end{table}

\paragraph{Score Range Adjustment is effective in mitigating bias and improving evaluation accuracy.}

Tables~\ref{tab:mtqe-score-scale} and \ref{tab:gecqe-score-scale} present the kurtosis and $r$ values for MTQE and GECQE under three score ranges: $\{1\text{--}5, 0\text{--}9, 1\text{--}100\}$.
Although the $1\text{--}5$ scale achieved the lowest kurtosis across most tasks and models, it did not necessarily produce the best evaluation performance.
This finding implies that while a low-kurtosis score range can sometimes enhance performance, it also carries the risk of producing random outputs.
However, the score ranges associated with the highest evaluation performances (e.g.,  the $1\text{--}100$ scale in Gemma with the MTQE task) also exhibited relatively lower kurtosis.
This pattern indicates that each model and task may have an optimal score range that both mitigates numerical bias and improves evaluation performance.
Moreover, the evaluation performances achieved with this optimal score range generally surpassed those obtained using the best temperature setting or the calibration method. For instance, in Gemma with the MTQE task,  the $1\text{--}100$ scale achieved 0.14 in $r$, outperforming the results in Figure \ref{fig:mtqe-temperature} and Table \ref{tab:calibration_mtqe}. 
Therefore, our experimental results suggest that adjusting the score range is the most promising strategy for mitigating numerical bias and enhancing evaluation performance.
While these results highlight score range adjustment as a comparatively effective mitigation strategy under current settings, it remains a heuristic and task-specific solution.
Identifying the optimal score range is nontrivial, and relying solely on range adjustment may not offer a generalizable fix.
To advance beyond symptom-level remedies, future work should also seek more fundamental approaches that directly address the causes of alignment-induced numerical bias.

\section{Discussion}
\subsection{Analysis of Input Features}
To detect inputs prone to bias, we analyze examples with strong bias.
Kurtosis captures dataset-level bias, but this analysis needs an example-level metric.
Thus, we use the \textbf{mode ratio}, the proportion of generated scores matching the data’s mode (the most frequently generated score in the dataset), as an indicator of bias strength.
This choice is motivated by our observation that in highly biased cases, a specific value appears frequently within 10 generations, often aligning with the data’s mode.
Assuming input complexity and edit extent affect bias, we compute the features in Table~\ref{tab:input-features} and calculate their Pearson correlation with the mode ratio.

Table~\ref{tab:pearson_mtqe} presents the results in the MTQE.
Except for Gemma, perplexity negatively correlates with the mode ratio, suggesting that more fluent sentences (lower perplexity) exhibit stronger numerical bias via frequent value repetition.

In GECQE, word overlap negatively correlates with the mode ratio across all models (Table~\ref{tab:pearson_gecqe}), indicating that more corrections (lower overlap) lead to stronger numerical bias.
Since GECQE emphasizes grammatical correctness, local lexical edits outweigh sentence-level fluency, possibly heightening the impact of word overlap.

\begin{table}[t]
    \centering
    \footnotesize
    \setlength\abovecaptionskip{2truemm}
    
    \renewcommand{\arraystretch}{0.9}
    \begin{tabular}{p{1.85cm}p{5.15cm}}
        \toprule
        Feature & Description \\
        \midrule
        Source Length & Number of words in the source sentence.\\
        Word Overlap & Overlap ratio between source and target.\\
        Source PPL & Perplexity of the source sentence.\\
        Target PPL & Perplexity of the target sentence.\\
        Overall PPL & Perplexity of both source and target.\\
        \bottomrule
    \end{tabular}
    \caption{List of features of input cases.}
    \label{tab:input-features}
\end{table}

\begin{table}[t]
    \centering
    \small
    \setlength\abovecaptionskip{2truemm}
    
    \renewcommand{\arraystretch}{0.9}
    \begin{tabular}{p{1cm}>{\raggedleft\arraybackslash}p{1cm}>{\raggedleft\arraybackslash}p{1.1cm}>{\raggedleft\arraybackslash}p{1cm}>{\raggedleft\arraybackslash}p{1cm}>{\raggedleft\arraybackslash}p{0.9cm}}
        \toprule
        Model & SLen & WOL & SPPL & TPPL & OPPL \\
        \midrule
        Gemma & 0.15 & \textbf{\textminus0.19} & \textminus0.02 & \textminus0.01 & 0.00 \\
        Mistral & 0.00 & \textminus0.08 & 0.00 & 0.00 & \textbf{\textminus0.19} \\
        Llama & \textminus0.14 & 0.00 & \textminus0.05 & \textminus0.12 & \textbf{\textminus0.21} \\
        Qwen & 0.12 & 0.08 & \textminus0.10 & \textminus0.17 & \textbf{\textminus0.31} \\
        \bottomrule
    \end{tabular}
    \caption{Pearson's correlation coefficient between input features and mode ratio in MTQE (En-De).}
    \label{tab:pearson_mtqe}
\end{table}

\begin{table}[t]
    \centering
    \small
    \setlength\abovecaptionskip{2truemm}
    
    \renewcommand{\arraystretch}{0.9}
    \begin{tabular}{p{1cm}>{\raggedleft\arraybackslash}p{1cm}>{\raggedleft\arraybackslash}p{1.1cm}>{\raggedleft\arraybackslash}p{1cm}>{\raggedleft\arraybackslash}p{1cm}>{\raggedleft\arraybackslash}p{0.9cm}}
        \toprule
        Model & SLen & WOL & SPPL & TPPL & OPPL \\
        \midrule
        Gemma & 0.05 & \textbf{\textminus0.15} & \textminus0.03 & \textminus0.03 & 0.00 \\
        Mistral & 0.13 & \textbf{\textminus0.34} & \textminus0.01 & \textminus0.03 & 0.02 \\
        Llama & 0.14 & \textbf{\textminus0.30} & \textminus0.02 & \textminus0.07 & \textminus0.04 \\
        Qwen & \textminus0.02 & \textbf{\textminus0.18} & 0.01 & \textminus0.06 & \textminus0.10 \\
        \bottomrule
    \end{tabular}
    \caption{Pearson's correlation coefficient between input features and mode ratio in GECQE (TMUGFM).}
    \label{tab:pearson_gecqe}
\end{table}

\subsection{Guidelines for Selecting Evaluator LLMs to Improve Accuracy and Reliability}  
Based on our results, we summarize key points for improving the accuracy and robustness of evaluations in LLM-as-a-judge.
First, pre-alignment models generally struggle with accuracy due to their limited instruction-following abilities.
In contrast, post-alignment models' stronger instruction-following abilities improve evaluation accuracy, making them the preferred choice as evaluators.

Second, while alignment induces numerical bias, the extent of its impact varies across models, highlighting the importance of careful model selection.
When gold scores are available, evaluation accuracy can be measured using correlation coefficients.
However, correlation alone is insufficient to assess robustness.
Combining correlation with kurtosis, which quantifies numerical bias, helps identify models that are accurate and unbiased across any data distribution.
When gold scores are unavailable, kurtosis serves as a useful proxy metric for model selection.
Our results indicate that models with high kurtosis, indicative of strong numerical bias, tend to have lower evaluation accuracy.

% Furthermore, adjusting the evaluation score range is effective in achieving the highest accuracy for the selected model.
% Since the optimal setting varies depending on the dataset and model, it is advisable to explore the best configuration for the model using validation data.
% Setting the optimal score range helps suppress numerical bias while enabling highly accurate evaluations.
Furthermore, adjusting the evaluation score range can improve accuracy for a given model, as it often reduces numerical bias.
However, the optimal setting varies by dataset and model, and thus requires careful tuning using validation data.
While this approach offers a practical way to improve evaluation stability, its effectiveness is task-specific, and score range should be treated as a configurable parameter rather than a fixed design choice.

\section{Conclusion}
This study clarified that post-alignment models exhibit an excessive concentration of scores, leading to numerical bias that compromises evaluation accuracy.
These findings suggest that the overconsistency induced by alignment may limit LLMs’ practical utility in quality evaluation tasks.
Future work should explore various alignment strengths and develop bias-mitigation strategies.

\section*{Limitations}
Our study has several limitations. 
(i) The public models used in our experiments do not disclose all of the methods and data used for alignment, making it impossible to analyze in depth the causes of the effects of alignment on data quality and methodological characteristics. 
Additionally, performing in-house alignment of pre-trained large language models is computationally expensive.
(ii) This study is limited to the target task using only numerical scores, and the effect of alignment on evaluations with natural language labels and rankings remains unresolved.
(iii) Both pre- and post-alignment models need to produce numerical outputs, which restricts the range of LLMs that can be used in the experiment.
(iv) Our mitigation methods, such as score range adjustment, are heuristic and task-specific. They represent a practical first step rather than a fundamental solution to numerical bias. Although effective in our setting, its generalizability remains uncertain, so further exploration in this direction is meaningful future work.

\bibliography{custom}

@inproceedings{fomicheva-etal-2022-mlqe,
    title = "{MLQE}-{PE}: A Multilingual Quality Estimation and Post-Editing Dataset",
    author = "Fomicheva, Marina  and
      Sun, Shuo  and
      Fonseca, Erick  and
      Zerva, Chrysoula  and
      Blain, Fr{\'e}d{\'e}ric  and
      Chaudhary, Vishrav  and
      Guzm{\'a}n, Francisco  and
      Lopatina, Nina  and
      Specia, Lucia  and
      Martins, Andr{\'e} F. T.",
    booktitle = "Proceedings of the Thirteenth Language Resources and Evaluation Conference",
    year = "2022",
    url = "https://aclanthology.org/2022.lrec-1.530",
    pages = "4963--4974",
}

@inproceedings{kocmi-federmann-2023-large,
    title = "Large Language Models Are State-of-the-Art Evaluators of Translation Quality",
    author = "Kocmi, Tom  and
      Federmann, Christian",
    booktitle = "Proceedings of the 24th Annual Conference of the European Association for Machine Translation",
    year = "2023",
    url = "https://aclanthology.org/2023.eamt-1.19",
    pages = "193--203",
}

@inproceedings{yoshimura-etal-2020-reference,
    title = "{SOME}: Reference-less Sub-Metrics Optimized for Manual Evaluations of Grammatical Error Correction",
    author = "Yoshimura, Ryoma  and
      Kaneko, Masahiro  and
      Kajiwara, Tomoyuki  and
      Komachi, Mamoru",
    booktitle = "Proceedings of the 28th International Conference on Computational Linguistics",
    month = dec,
    year = "2020",
    address = "Barcelona, Spain (Online)",
    publisher = "International Committee on Computational Linguistics",
    url = "https://aclanthology.org/2020.coling-main.573",
    doi = "10.18653/v1/2020.coling-main.573",
    pages = "6516--6522",
}

@inproceedings{suzuki-etal-2022-construction,
    title = "Construction of a Quality Estimation Dataset for Automatic Evaluation of {J}apanese Grammatical Error Correction",
    author = "Suzuki, Daisuke  and
      Takahashi, Yujin  and
      Yamashita, Ikumi  and
      Aida, Taichi  and
      Hirasawa, Tosho  and
      Nakatsuji, Michitaka  and
      Mita, Masato  and
      Komachi, Mamoru",
    booktitle = "Proceedings of the Thirteenth Language Resources and Evaluation Conference",
    month = jun,
    year = "2022",
    address = "Marseille, France",
    publisher = "European Language Resources Association",
    url = "https://aclanthology.org/2022.lrec-1.596",
    pages = "5565--5572",
}

@inproceedings{shardlow-etal-2020-complex,
    title = "{C}omp{L}ex {---} A New Corpus for Lexical Complexity Prediction from {L}ikert {S}cale Data",
    author = "Shardlow, Matthew  and
      Cooper, Michael  and
      Zampieri, Marcos",
    booktitle = "Proceedings of the 1st Workshop on Tools and Resources to Empower People with REAding DIfficulties (READI)",
    month = may,
    year = "2020",
    address = "Marseille, France",
    publisher = "European Language Resources Association",
    url = "https://aclanthology.org/2020.readi-1.9",
    pages = "57--62",
    language = "English",
    ISBN = "979-10-95546-45-0",
}

@inproceedings{papineni-etal-2002-bleu,
    title = "{B}leu: a Method for Automatic Evaluation of Machine Translation",
    author = "Papineni, Kishore  and
      Roukos, Salim  and
      Ward, Todd  and
      Zhu, Wei-Jing",
    editor = "Isabelle, Pierre  and
      Charniak, Eugene  and
      Lin, Dekang",
    booktitle = "Proceedings of the 40th Annual Meeting of the Association for Computational Linguistics",
    month = jul,
    year = "2002",
    address = "Philadelphia, Pennsylvania, USA",
    publisher = "Association for Computational Linguistics",
    url = "https://aclanthology.org/P02-1040",
    doi = "10.3115/1073083.1073135",
    pages = "311--318",
}

@InProceedings{pmlr-v202-santurkar23a,
  title = 	 {Whose Opinions Do Language Models Reflect?},
  author =       {Santurkar, Shibani and Durmus, Esin and Ladhak, Faisal and Lee, Cinoo and Liang, Percy and Hashimoto, Tatsunori},
  booktitle = 	 {Proceedings of the 40th International Conference on Machine Learning},
  pages = 	 {29971--30004},
  year = 	 {2023},
  editor = 	 {Krause, Andreas and Brunskill, Emma and Cho, Kyunghyun and Engelhardt, Barbara and Sabato, Sivan and Scarlett, Jonathan},
  volume = 	 {202},
  series = 	 {Proceedings of Machine Learning Research},
  month = 	 {23--29 Jul},
  publisher =    {PMLR},
  url = 	 {https://proceedings.mlr.press/v202/santurkar23a.html}
}

@misc{mohammadi2024creativityleftchatprice,
      title={Creativity Has Left the Chat: The Price of Debiasing Language Models}, 
      author={Behnam Mohammadi},
      year={2024},
      eprint={2406.05587},
      archivePrefix={arXiv},
      primaryClass={cs.CL},
      url={https://arxiv.org/abs/2406.05587}, 
}

@inproceedings{NEURIPS2022-b1efde53,
     author = {Ouyang, Long and Wu, Jeffrey and Jiang, Xu and Almeida, Diogo and Wainwright, Carroll and Mishkin, Pamela and Zhang, Chong and Agarwal, Sandhini and Slama, Katarina and Ray, Alex and Schulman, John and Hilton, Jacob and Kelton, Fraser and Miller, Luke and Simens, Maddie and Askell, Amanda and Welinder, Peter and Christiano, Paul F and Leike, Jan and Lowe, Ryan},
     booktitle = {Advances in Neural Information Processing Systems},
     editor = {S. Koyejo and S. Mohamed and A. Agarwal and D. Belgrave and K. Cho and A. Oh},
     pages = {27730--27744},
     publisher = {Curran Associates, Inc.},
     title = {Training language models to follow instructions with human feedback},
     url = {https://proceedings.neurips.cc/paper\_files/paper/2022/file/b1efde53be364a73914f58805a001731-Paper-Conference.pdf},
     volume = {35},
     year = {2022}
}

@inproceedings{ohi-etal-2024-likelihood,
    title = "Likelihood-based Mitigation of Evaluation Bias in Large Language Models",
    author = "Ohi, Masanari  and
      Kaneko, Masahiro  and
      Koike, Ryuto  and
      Loem, Mengsay  and
      Okazaki, Naoaki",
    editor = "Ku, Lun-Wei  and
      Martins, Andre  and
      Srikumar, Vivek",
    booktitle = "Findings of the Association for Computational Linguistics ACL 2024",
    month = aug,
    year = "2024",
    address = "Bangkok, Thailand and virtual meeting",
    publisher = "Association for Computational Linguistics",
    url = "https://aclanthology.org/2024.findings-acl.193",
    doi = "10.18653/v1/2024.findings-acl.193",
    pages = "3237--3245",
}

@inproceedings{enomoto-etal-2024-tmu,
    title = "{TMU}-{HIT} at {MLSP} 2024: How Well Can {GPT}-4 Tackle Multilingual Lexical Simplification?",
    author = "Enomoto, Taisei  and
      Kim, Hwichan  and
      Hirasawa, Tosho  and
      Nagai, Yoshinari  and
      Sato, Ayako  and
      Nakajima, Kyotaro  and
      Komachi, Mamoru",
    booktitle = "Proceedings of the 19th Workshop on Innovative Use of NLP for Building Educational Applications (BEA 2024)",
    year = "2024",
    url = "https://aclanthology.org/2024.bea-1.52",
    pages = "590--598",
}

@misc{stureborg2024largelanguagemodelsinconsistent,
      title={Large Language Models are Inconsistent and Biased Evaluators}, 
      author={Rickard Stureborg and Dimitris Alikaniotis and Yoshi Suhara},
      year={2024},
      eprint={2405.01724},
      archivePrefix={arXiv},
      primaryClass={cs.CL},
      url={https://arxiv.org/abs/2405.01724}, 
}

@inproceedings{koyama-etal-2020-construction,
    title = "Construction of an Evaluation Corpus for Grammatical Error Correction for Learners of {J}apanese as a Second Language",
    author = "Koyama, Aomi  and
      Kiyuna, Tomoshige  and
      Kobayashi, Kenji  and
      Arai, Mio  and
      Komachi, Mamoru",
    editor = "Calzolari, Nicoletta  and
      B{\'e}chet, Fr{\'e}d{\'e}ric  and
      Blache, Philippe  and
      Choukri, Khalid  and
      Cieri, Christopher  and
      Declerck, Thierry  and
      Goggi, Sara  and
      Isahara, Hitoshi  and
      Maegaard, Bente  and
      Mariani, Joseph  and
      Mazo, H{\'e}l{\`e}ne  and
      Moreno, Asuncion  and
      Odijk, Jan  and
      Piperidis, Stelios",
    booktitle = "Proceedings of the Twelfth Language Resources and Evaluation Conference",
    month = may,
    year = "2020",
    address = "Marseille, France",
    publisher = "European Language Resources Association",
    url = "https://aclanthology.org/2020.lrec-1.26",
    pages = "204--211",
    language = "English",
    ISBN = "979-10-95546-34-4",
}

@inproceedings{kocmi-federmann-2023-gemba,
    title = "{GEMBA}-{MQM}: Detecting Translation Quality Error Spans with {GPT}-4",
    author = "Kocmi, Tom  and
      Federmann, Christian",
    editor = "Koehn, Philipp  and
      Haddow, Barry  and
      Kocmi, Tom  and
      Monz, Christof",
    booktitle = "Proceedings of the Eighth Conference on Machine Translation",
    month = dec,
    year = "2023",
    address = "Singapore",
    publisher = "Association for Computational Linguistics",
    url = "https://aclanthology.org/2023.wmt-1.64",
    doi = "10.18653/v1/2023.wmt-1.64",
    pages = "768--775",
}

@inproceedings{lin-2004-rouge,
    title = "{ROUGE}: A Package for Automatic Evaluation of Summaries",
    author = "Lin, Chin-Yew",
    booktitle = "Text Summarization Branches Out",
    month = jul,
    year = "2004",
    address = "Barcelona, Spain",
    publisher = "Association for Computational Linguistics",
    url = "https://aclanthology.org/W04-1013",
    pages = "74--81",
}

@inproceedings{kobayashi-etal-2024-large,
    title = "Large Language Models Are State-of-the-Art Evaluator for Grammatical Error Correction",
    author = "Kobayashi, Masamune  and
      Mita, Masato  and
      Komachi, Mamoru",
    editor = {Kochmar, Ekaterina  and
      Bexte, Marie  and
      Burstein, Jill  and
      Horbach, Andrea  and
      Laarmann-Quante, Ronja  and
      Tack, Ana{\"\i}s  and
      Yaneva, Victoria  and
      Yuan, Zheng},
    booktitle = "Proceedings of the 19th Workshop on Innovative Use of NLP for Building Educational Applications (BEA 2024)",
    month = jun,
    year = "2024",
    address = "Mexico City, Mexico",
    publisher = "Association for Computational Linguistics",
    url = "https://aclanthology.org/2024.bea-1.6",
    pages = "68--77",
}

@inproceedings{specia-etal-2020-findings-wmt,
    title = "Findings of the {WMT} 2020 Shared Task on Quality Estimation",
    author = "Specia, Lucia  and
      Blain, Fr{\'e}d{\'e}ric  and
      Fomicheva, Marina  and
      Fonseca, Erick  and
      Chaudhary, Vishrav  and
      Guzm{\'a}n, Francisco  and
      Martins, Andr{\'e} F. T.",
    editor = {Barrault, Lo{\"\i}c  and
      Bojar, Ond{\v{r}}ej  and
      Bougares, Fethi  and
      Chatterjee, Rajen  and
      Costa-juss{\`a}, Marta R.  and
      Federmann, Christian  and
      Fishel, Mark  and
      Fraser, Alexander  and
      Graham, Yvette  and
      Guzman, Paco  and
      Haddow, Barry  and
      Huck, Matthias  and
      Yepes, Antonio Jimeno  and
      Koehn, Philipp  and
      Martins, Andr{\'e}  and
      Morishita, Makoto  and
      Monz, Christof  and
      Nagata, Masaaki  and
      Nakazawa, Toshiaki  and
      Negri, Matteo},
    booktitle = "Proceedings of the Fifth Conference on Machine Translation",
    month = nov,
    year = "2020",
    address = "Online",
    publisher = "Association for Computational Linguistics",
    url = "https://aclanthology.org/2020.wmt-1.79",
    pages = "743--764",
}

@inproceedings{
    zheng2023judging,
    title={Judging {LLM}-as-a-Judge with {MT}-Bench and Chatbot Arena},
    author={Lianmin Zheng and Wei-Lin Chiang and Ying Sheng and Siyuan Zhuang and Zhanghao Wu and Yonghao Zhuang and Zi Lin and Zhuohan Li and Dacheng Li and Eric Xing and Hao Zhang and Joseph E. Gonzalez and Ion Stoica},
    booktitle={Thirty-seventh Conference on Neural Information Processing Systems Datasets and Benchmarks Track},
    year={2023},
    url={https://openreview.net/forum?id=uccHPGDlao}
}

@inproceedings{chiang-lee-2023-large,
    title = "Can Large Language Models Be an Alternative to Human Evaluations?",
    author = "Chiang, Cheng-Han  and
      Lee, Hung-yi",
    editor = "Rogers, Anna  and
      Boyd-Graber, Jordan  and
      Okazaki, Naoaki",
    booktitle = "Proceedings of the 61st Annual Meeting of the Association for Computational Linguistics (Volume 1: Long Papers)",
    month = jul,
    year = "2023",
    address = "Toronto, Canada",
    publisher = "Association for Computational Linguistics",
    url = "https://aclanthology.org/2023.acl-long.870",
    doi = "10.18653/v1/2023.acl-long.870",
    pages = "15607--15631",
}

@inproceedings{sato-etal-2024-tmu,
    title = "{TMU}-{HIT}{'}s Submission for the {WMT}24 Quality Estimation Shared Task: Is {GPT}-4 a Good Evaluator for Machine Translation?",
    author = "Sato, Ayako  and
      Nakajima, Kyotaro  and
      Kim, Hwichan  and
      Chen, Zhousi  and
      Komachi, Mamoru",
    editor = "Haddow, Barry  and
      Kocmi, Tom  and
      Koehn, Philipp  and
      Monz, Christof",
    booktitle = "Proceedings of the Ninth Conference on Machine Translation",
    month = nov,
    year = "2024",
    address = "Miami, Florida, USA",
    publisher = "Association for Computational Linguistics",
    url = "https://aclanthology.org/2024.wmt-1.38",
    doi = "10.18653/v1/2024.wmt-1.38",
    pages = "529--534",
}

@inproceedings{
rafailov2023direct,
title={Direct Preference Optimization: Your Language Model is Secretly a Reward Model},
author={Rafael Rafailov and Archit Sharma and Eric Mitchell and Christopher D Manning and Stefano Ermon and Chelsea Finn},
booktitle={Thirty-seventh Conference on Neural Information Processing Systems},
year={2023},
url={https://openreview.net/forum?id=HPuSIXJaa9}
}

@inproceedings{
wei2022finetuned,
title={Finetuned Language Models are Zero-Shot Learners},
author={Jason Wei and Maarten Bosma and Vincent Zhao and Kelvin Guu and Adams Wei Yu and Brian Lester and Nan Du and Andrew M. Dai and Quoc V Le},
booktitle={International Conference on Learning Representations},
year={2022},
url={https://openreview.net/forum?id=gEZrGCozdqR}
}

@inproceedings{jiang-etal-2023-generative,
    title = "Generative Calibration for In-context Learning",
    author = "Jiang, Zhongtao  and
      Zhang, Yuanzhe  and
      Liu, Cao  and
      Zhao, Jun  and
      Liu, Kang",
    editor = "Bouamor, Houda  and
      Pino, Juan  and
      Bali, Kalika",
    booktitle = "Findings of the Association for Computational Linguistics: EMNLP 2023",
    month = dec,
    year = "2023",
    address = "Singapore",
    publisher = "Association for Computational Linguistics",
    url = "https://aclanthology.org/2023.findings-emnlp.152/",
    doi = "10.18653/v1/2023.findings-emnlp.152",
    pages = "2312--2333",
}

\appendix

\section{Implementation Details}
\paragraph{Models}
We conducted experiments with Gemma\footnote{\scriptsize \url{https://huggingface.co/google/gemma-7b}}\fnsep\footnote{\scriptsize \url{https://huggingface.co/google/gemma-7b-it}}, Mistral\footnote{\scriptsize \url{https://huggingface.co/mistralai/Mistral-7B-v0.1}}\fnsep\footnote{\scriptsize \url{https://huggingface.co/mistralai/Mistral-7B-Instruct-v0.1}}, Llama-3\footnote{\scriptsize \url{https://huggingface.co/meta-llama/Meta-Llama-3-8B}}\fnsep\footnote{\scriptsize \url{https://huggingface.co/meta-llama/Meta-Llama-3-8B-Instruct}} and Qwen-2\footnote{\scriptsize \url{https://huggingface.co/Qwen/Qwen2-7B}}\fnsep\footnote{\scriptsize \url{https://huggingface.co/Qwen/Qwen2-7B-Instruct}} from huggingface and used a single Quadro RTX $8000$ in the all experiments.
Gemma and Llama-3 are published by their own licenses\footnote{\scriptsize \url{https://ai.google.dev/gemma/terms}}\fnsep\footnote{\scriptsize \url{https://www.llama.com/llama3/license/}}.
Mistral and Qwen-2 are published by the Apache 2.0 license.

\paragraph{Data}
In MTQE, we used the test data from WMT QE 2020\footnote{\scriptsize \url{https://github.com/WMT-QE-Task/wmt-qe-2022-data/tree/main/train-dev_data/task1_da/train}}, obtained from the training data of WMT QE 2022.
For calibration, the WMT QE 2021 dev data was used to estimate the true data distribution $q(y)$, while the WMT QE 2021 train data was used to estimate the beta distribution.
Each language pair contains $1,000$ pairs of source texts and translations.
In GECQE, we used the TMU-GFM-Dataset\footnote{\scriptsize \url{https://huggingface.co/datasets/tmu-nlp/tmu_gfm_dataset}} and autoJQE\footnote{\scriptsize \url{https://github.com/tmu-nlp/autoJQE/tree/main}}.
Since these datasets are not pre-split into train, dev, or test sets, we allocated 80\% of the data as test data for experiments, considering the possibility of using some data as dev data or as a pool for few-shot examples.
As a result, the dataset sizes were $3,377$ for TMUGFM, $3,773$ for autoJQE-FLUTEC, and $3,513$ for autoJQE-TECJL.
In LCP, we used MLSP2024\footnote{\scriptsize \url{https://huggingface.co/datasets/MLSP2024/MLSP2024}}.
We use nine languages except for Portuguese, for which data was partially corrupted.
The data size for each language is shown in Table \ref{tab:mlsp2024}.

\begin{table}
    \centering
    \begin{tabular}{lc}
    \toprule
        Language & Data size \\
    \midrule
        Catalan (Ca) & 445 \\
        English (En)& 570 \\
        Filipino (Fil)& 570 \\
        French (Fr)& 569 \\
        German (De) & 570 \\
        Italian (It) & 570 \\
        Japanese (Ja) & 570 \\
        Sinhala (Es) & 600 \\
        Spanish (Es) & 593 \\
    \bottomrule
    \end{tabular}
    \caption{The data size for each language in MLSP2024.}
    \label{tab:mlsp2024}
\end{table}

\section{Prompt Template}
\label{sec:appendix-prompt}
We define the prompt template for the GECQE task as shown in Table \ref{tab:prompt-template-gecqe}, and for the LCP task as shown in Table \ref{tab:prompt-template-lcp}.

\begin{table}[h]
    \centering
    \small
    \setlength\abovecaptionskip{2truemm}
    
    \begin{tabular}{p{7.25cm}}
        \toprule
        Please analyze the given original and corrected sentences and output a grammatical error correction quality score on a integer scale ranging from \{\{min score\}\} to \{\{max score\}\}.\\
        A score close to \{\{min score\}\} indicates a low quality correction , while a score close to \{\{max score\}\} indicates a high quality correction.\\
        Do not provide any explanations or text apart from the score.\\
        
        Original sentence: \{\{$org_i$\}\}\\
        Corrected sentence: \{\{$cor_i$\}\}\\
        Score:\\
        \bottomrule
    \end{tabular}
    \caption{Prompt Template for GECQE.}
    \label{tab:prompt-template-gecqe}
\end{table}

\begin{table}[h]
    \centering
    \small
    \setlength\abovecaptionskip{2truemm}
    
    \begin{tabular}{p{7.25cm}}
        \toprule
        Please analyze the given sentence and word included in the sentence and output a complexity score on a integer scale ranging from \{\{min score\}\} to \{\{max score\}\}.\\
        The complexity score should be evaluated based on the difficulty of the word included in the sentence.\\
        A score closer to \{\{min score\}\} indicates that the word is easy, while a score closer to \{\{max score\}\} indicates that the word is difficult.\\
        Do not provide any explanations or text apart from the score.\\
        
        Sentence: \{\{$sent_i$\}\}\\
        Word: \{\{$word_i$\}\}\\
        Score:\\
        \bottomrule
    \end{tabular}
    \caption{Prompt Template for LCP.}
    \label{tab:prompt-template-lcp}
\end{table}

\section{Supplementary Experimental Results}
\label{sec:appendix-result}
Figure \ref{fig:lcp_0-9} and Table \ref{tab:lcp_0-9} present the LCP task results for RQ1.
In many languages, the score distributions tend to become skewed after alignment, a phenomenon also observed in the MTQE and GECQE tasks.
In addition, the MTQE results for other languages and the GECQE results for FLUTEC are shown in Figures \ref{fig:mtqe_0-9_appendix} and \ref{fig:gecqe_0-9_appendix}, respectively.
Furthermore, Figure \ref{fig:scatter_0-9} displays scatter plots for the GECQE (TMUGFM) and for the LCP (English).

{\setlength{\tabcolsep}{4.5pt}
\begin{table}[t]
    \centering
    \small
    \setlength\abovecaptionskip{2truemm}
    
    \renewcommand{\arraystretch}{0.9}
    \begin{tabular}{llrrrcc}
    \toprule
        & & \multicolumn{3}{c}{Kurtosis} & \multicolumn{2}{c}{$r$} \\
        \cmidrule(lr){3-5}
        \cmidrule(lr){6-7}
        \multicolumn{1}{c}{Lang} & \multicolumn{1}{c}{Model} & \multicolumn{1}{c}{gold} & \multicolumn{1}{c}{pre} & \multicolumn{1}{c}{post} & \multicolumn{1}{c}{pre} & \multicolumn{1}{c}{post} \\
    \midrule
        \multirow{4}{*}{Catalan} & Gemma & \textminus0.50 & \textminus0.21 & \textbf{\textminus0.18} & 0.06\hphantom{*} & \textbf{0.16$^{*}$} \\
        & Mistral & \textminus0.50 & \textbf{\textminus0.04} & 2.60 & 0.02\hphantom{*} & \textbf{0.17$^{*}$} \\
        & Llama & \textminus0.50 & \textbf{1.17} & 11.92 \textbf{}& 0.14$^{*}$ & \textbf{0.19$^{*}$} \\
        & Qwen & \textminus0.50 & \textbf{0.08} & 6.70 & 0.26$^{*}$ & \textbf{0.28$^{*}$} \\
    \midrule
        \multirow{4}{*}{English} & Gemma & 1.36 & \textbf{\textminus0.20} & \textminus0.42 & 0.19$^{*}$ & \textbf{0.35$^{*}$} \\
        & Mistral & 1.36 & \textbf{0.02} & 0.58 & 0.14$^{*}$ & \textbf{0.36$^{*}$} \\
        & Llama & 1.36 & \textbf{\textminus0.07} & \textminus0.89 & 0.43$^{*}$ & \textbf{0.64$^{*}$} \\
        & Qwen & 1.36 & \textbf{0.27} & \textminus0.56 & 0.62$^{*}$ & \textbf{0.68$^{*}$} \\
    \midrule
        \multirow{4}{*}{Filipino} & Gemma & 0.00 & \textbf{\textminus0.12} & 1.42 & 0.12$^{*}$ & \textbf{0.32$^{*}$} \\
        & Mistral & 0.00 & \textbf{0.01} & 1.17 & 0.04\hphantom{*} & \textbf{0.23$^{*}$} \\
        & Llama & 0.00 & \textbf{\textminus0.31} & 2.47 & 0.25$^{*}$ & \textbf{0.42$^{*}$} \\
        & Qwen & 0.00 & \textbf{\textminus0.37} & \textminus1.10 & 0.37$^{*}$ & \textbf{0.39$^{*}$} \\
    \midrule
        \multirow{4}{*}{French} & Gemma & \textminus0.58 & \textbf{\textminus0.32} & 0.57 & 0.07\hphantom{*} & \textbf{0.33$^{*}$} \\
        & Mistral & \textminus0.58 & \textbf{\textminus0.48} & 2.85 & \textbf{0.08\hphantom{*}} & 0.06\hphantom{*} \\
        & Llama & \textminus0.58 & \textbf{0.23} & 2.83 & 0.24$^{*}$ & \textbf{0.32$^{*}$} \\
        & Qwen & \textminus0.58 & \textbf{\textminus0.71} & \textminus0.87 & 0.28$^{*}$ & \textbf{0.33$^{*}$} \\
    \midrule
        \multirow{4}{*}{German}& Gemma & \textminus0.33 & \textbf{\textminus0.26} & 2.83 & 0.11\hphantom{*} & \textbf{0.30$^{*}$} \\
        & Mistral & \textminus0.33 & \textbf{0.27} & 6.25 & 0.10\hphantom{*} & \textbf{0.14$^{*}$} \\
        & Llama & \textminus0.33 & \textbf{\textminus0.60} & 1.88 & 0.27$^{*}$ & \textbf{0.34$^{*}$} \\
        & Qwen & \textminus0.33 & \textbf{0.83} & 2.64 & 0.28$^{*}$ & \textbf{0.31$^{*}$} \\
    \midrule
        \multirow{4}{*}{Italian} & Gemma & 0.29 & \textbf{\textminus0.16} & 0.22 & 0.10$^{*}$ & \textbf{0.28$^{*}$} \\
        & Mistral & 0.29 & \textbf{0.31} & 7.29 & 0.09$^{*}$ & \textbf{0.22$^{*}$} \\
        & Llama & 0.29 & \textbf{\textminus0.16} & 5.95 & 0.19$^{*}$ & \textbf{0.27$^{*}$} \\
        & Qwen & 0.29 & \textbf{\textminus0.10} & 0.26 & 0.31$^{*}$ & \textbf{0.36$^{*}$} \\
    \midrule
        \multirow{4}{*}{Japanese} & Gemma & \textminus0.55 & \textbf{0.33} & 1.53 & 0.08\hphantom{*} & \textbf{0.26$^{*}$} \\
        & Mistral & \textminus0.55 & \textbf{\textminus0.25} & 1.37 & 0.11\hphantom{*} & \textbf{0.12$^{*}$} \\
        & Llama & \textminus0.55 & \textbf{\textminus0.33} & 0.37 & 0.22$^{*}$ & \textbf{0.42$^{*}$} \\
        & Qwen & \textminus0.55 & \textminus0.57 & \textbf{\textminus0.36} & 0.31$^{*}$ & \textbf{0.36}$^{*}$ \\
    \midrule
        \multirow{4}{*}{Sinhala} & Gemma & 3.43 & \textbf{0.02} & 1.73 & 0.02$^{*}$ & \textbf{0.02$^{*}$} \\
        & Mistral & 3.43 & \textbf{\textminus0.18} & 2.62 & 0.00$^{*}$ & \textbf{0.00$^{*}$} \\
        & Llama & 3.43 & \textbf{2.22} & 18.41 & 0.06$^{*}$ & \textbf{0.08$^{*}$} \\
        & Qwen & 3.43 & \textbf{0.22} & 2.73 & 0.03$^{*}$ & \textbf{0.09$^{*}$} \\
    \midrule
    \multirow{4}{*}{Spanish} & Gemma & 0.50 & \textbf{0.11} & 0.59 & 0.22$^{*}$ & \textbf{0.40$^{*}$} \\
        & Mistral & 0.50 & \textbf{0.02} & 4.04 & 0.14$^{*}$ & \textbf{0.35$^{*}$} \\
        & Llama & 0.50 & \textminus0.66 & \textbf{\textminus0.26} & 0.35$^{*}$ & \textbf{0.48$^{*}$} \\
        & Qwen & 0.50 & \textbf{\textminus0.88} & \textminus1.31 & 0.45$^{*}$ & \textbf{0.53$^{*}$} \\
    \bottomrule
    \end{tabular}
    \caption{Kurtosis of LCP score distributions and Pearson correlation coefficient ($r$). For each language pair and model, the \underline{smallest absolute kurtosis} and the \underline{highest $r$} are highlighted in \textbf{bold}. $^{*}$ indicate statistically significant correlations ($p<0.01$).}
    \label{tab:lcp_0-9}
\end{table}}

\begin{figure}[t]
    \setlength\abovecaptionskip{2truemm}
    \centering
    \renewcommand{\arraystretch}{0.9}
    \begin{minipage}[b]{\linewidth}
        \centering
        \includegraphics[width=\linewidth]{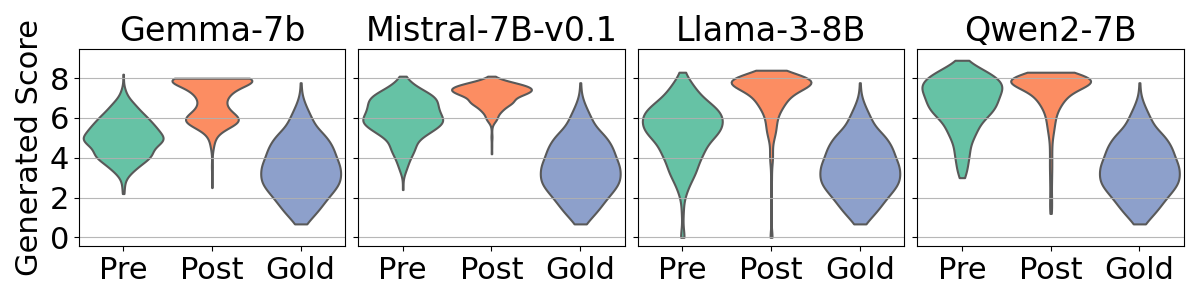}
        \subcaption{Calalan}
    \end{minipage}
    \begin{minipage}[b]{\linewidth}
        \centering
        \includegraphics[width=\linewidth]{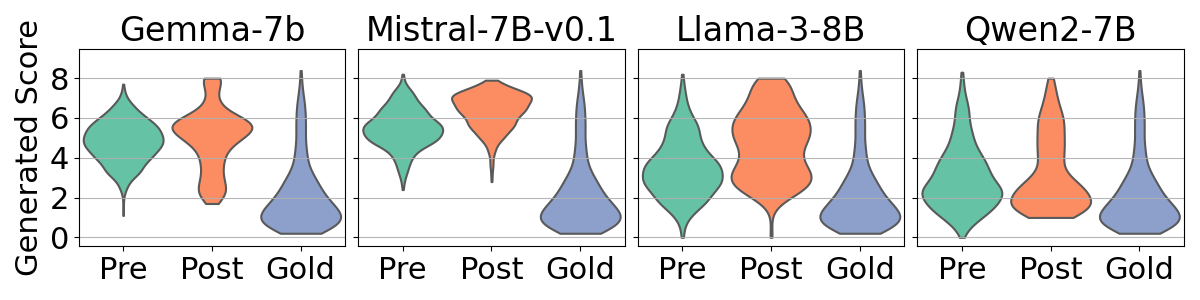}
        \subcaption{English}
    \end{minipage}
    \begin{minipage}[b]{\linewidth}
        \centering
        \includegraphics[width=\linewidth]{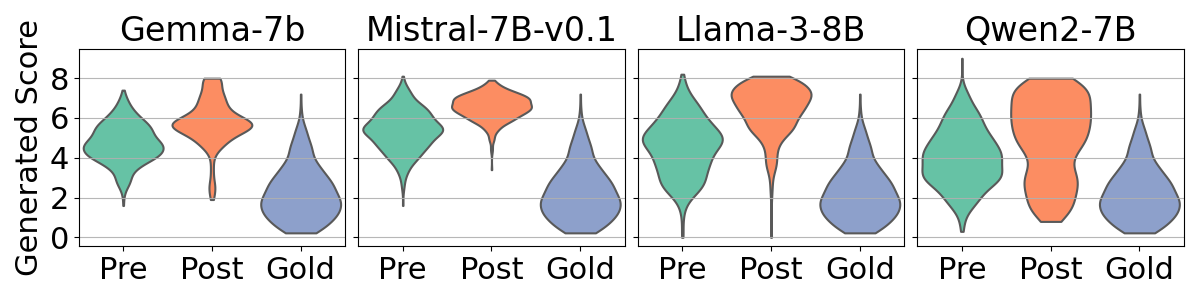}
        \subcaption{Filipino}
    \end{minipage}
    \begin{minipage}[b]{\linewidth}
        \centering
        \includegraphics[width=\linewidth]{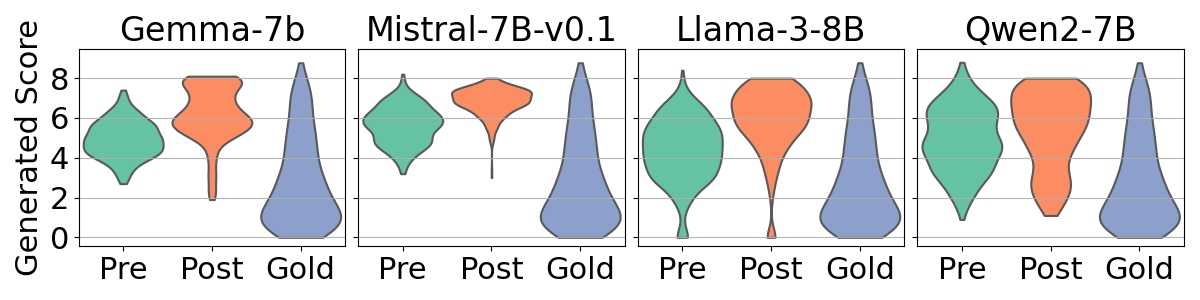}
        \subcaption{French}
    \end{minipage}
    \begin{minipage}[b]{\linewidth}
        \centering
        \includegraphics[width=\linewidth]{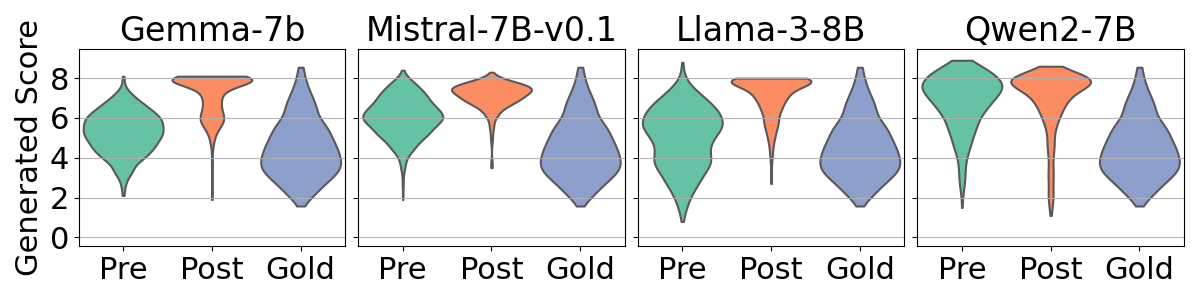}
        \subcaption{German}
    \end{minipage}
    \begin{minipage}[b]{\linewidth}
        \centering
        \includegraphics[width=\linewidth]{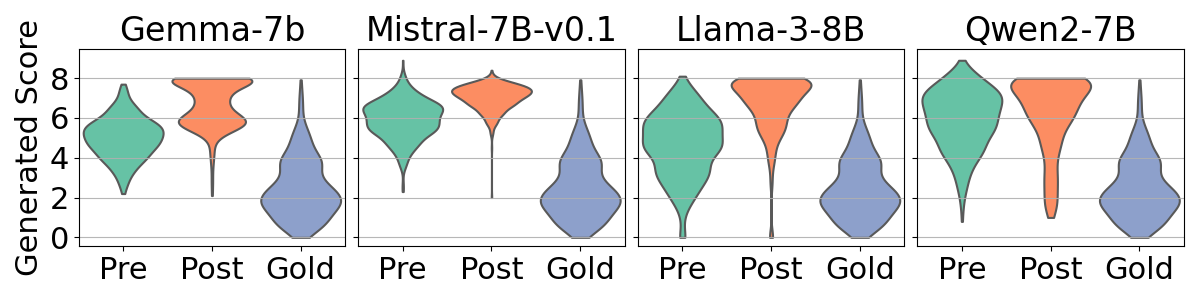}
        \subcaption{Italian}
    \end{minipage}
    \begin{minipage}[b]{\linewidth}
        \centering
        \includegraphics[width=\linewidth]{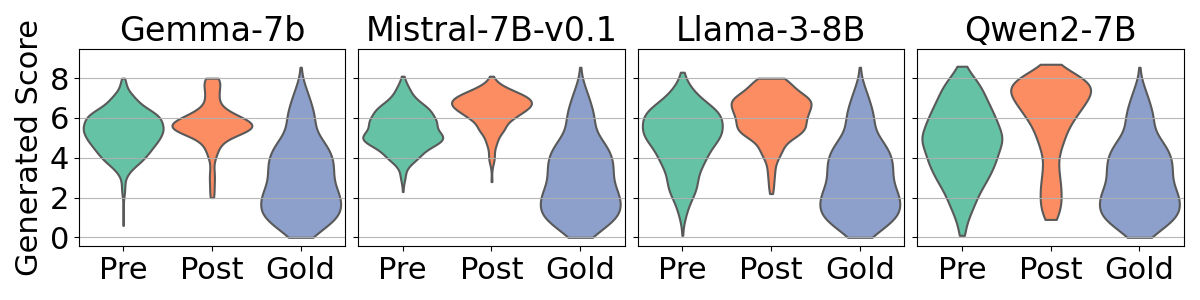}
        \subcaption{Japanese}
    \end{minipage}
    \begin{minipage}[b]{\linewidth}
        \centering
        \includegraphics[width=\linewidth]{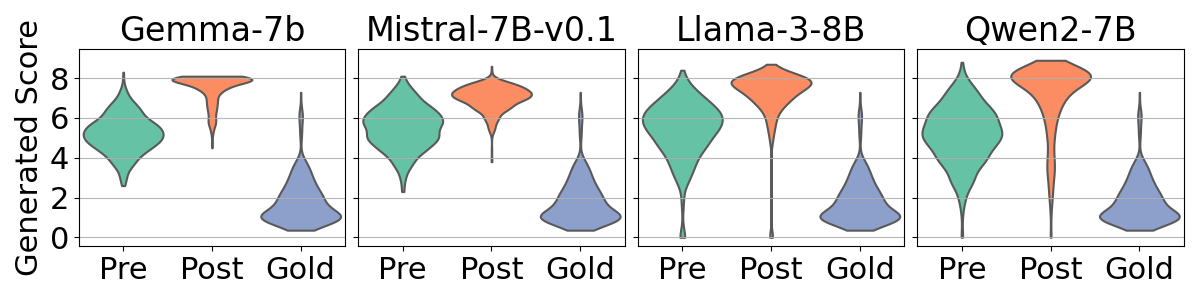}
        \subcaption{Sinhala}
    \end{minipage}
    \begin{minipage}[b]{\linewidth}
        \centering
        \includegraphics[width=\linewidth]{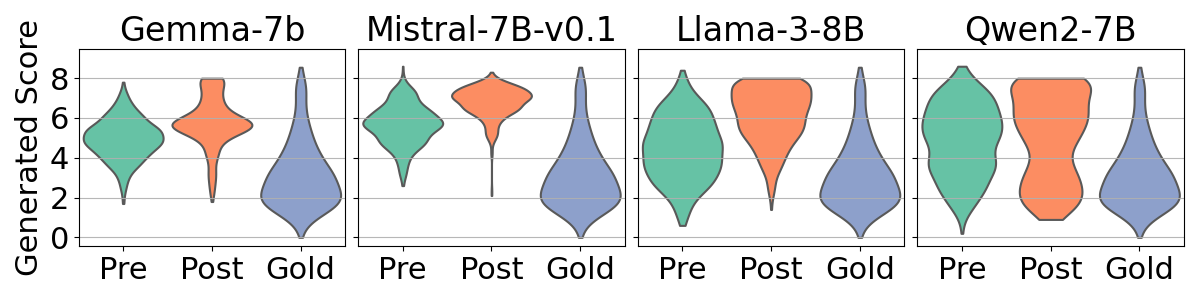}
        \subcaption{Spanish}
    \end{minipage}
    \caption{LCP score distribution.}
    \label{fig:lcp_0-9}
\end{figure}

\begin{figure}[t]
    \setlength\abovecaptionskip{2truemm}
    \centering
    \begin{minipage}[b]{\linewidth}
        \centering
        \includegraphics[width=\linewidth]{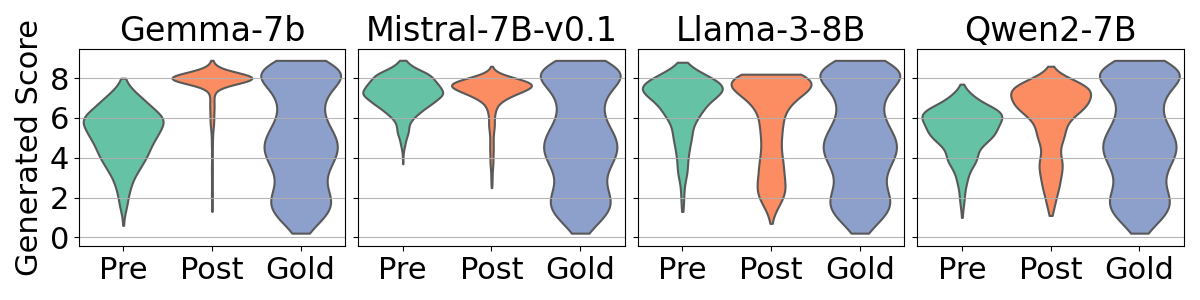}
        \subcaption{Estonian to English (Et-En)}
    \end{minipage}
    \begin{minipage}[b]{\linewidth}
        \centering
        \includegraphics[width=\linewidth]{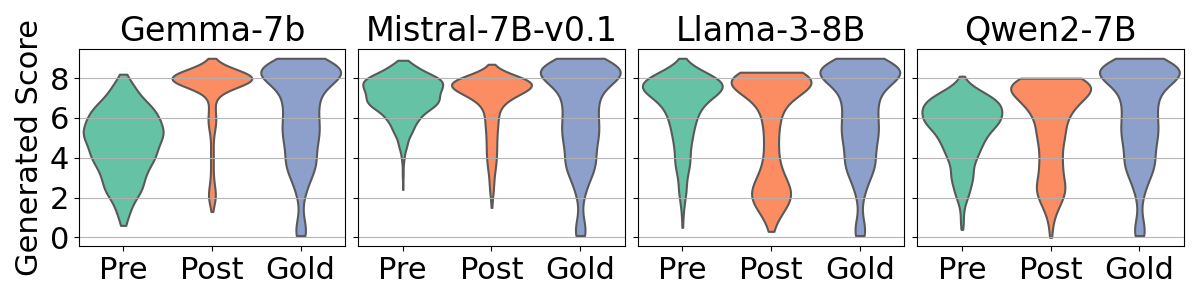}
        \subcaption{Romanian to English (Ro-En)}
    \end{minipage}
    \begin{minipage}[b]{\linewidth}
        \centering
        \includegraphics[width=\linewidth]{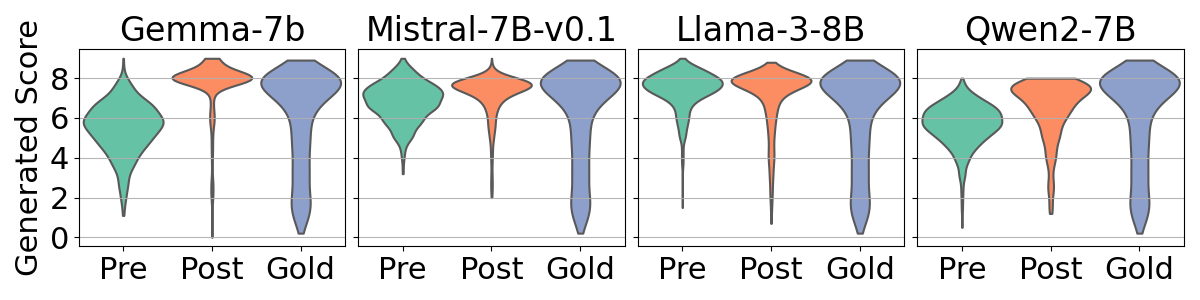}
        \subcaption{Russian to English (Ru-En)}
    \end{minipage}
    \caption{MTQE score distribution.}
    \label{fig:mtqe_0-9_appendix}
\end{figure}

\begin{figure}[t]
    \setlength\abovecaptionskip{2truemm}
    \centering
        \includegraphics[width=\linewidth]{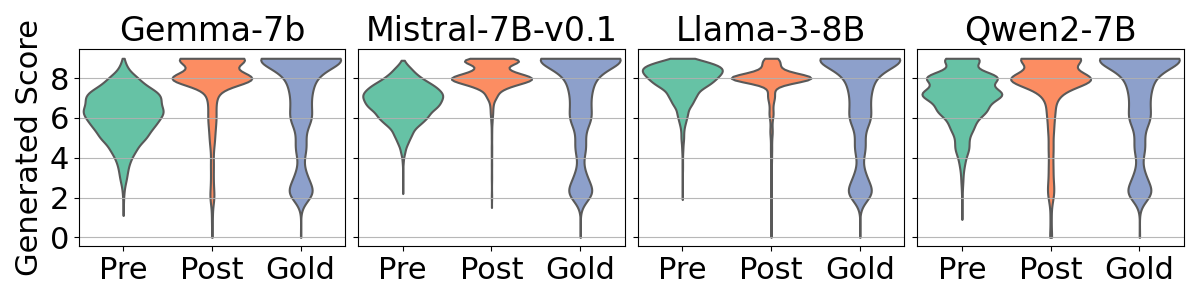}
    \caption{Distribution of GECQE scores in FLUTEC (Japanese).}
    \label{fig:gecqe_0-9_appendix}
\end{figure}

\begin{figure}[t]
    \setlength\abovecaptionskip{2truemm}
    \centering
    \begin{minipage}[b]{\linewidth}
        \centering
        \includegraphics[width=\linewidth]{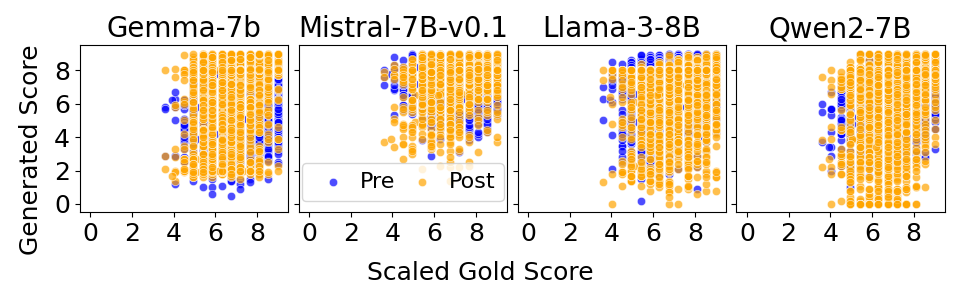}
        \subcaption{GECQE (TMUGFM; English)}
    \end{minipage}
    \begin{minipage}[b]{\linewidth}
        \centering
        \includegraphics[width=\linewidth]{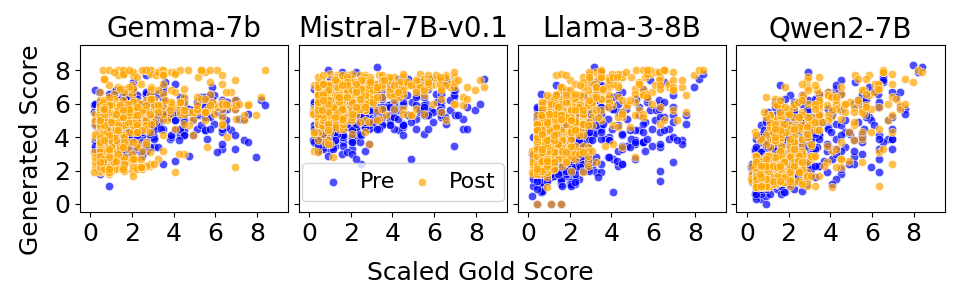}
        \subcaption{LCP (English)}
    \end{minipage}
    \caption{Scatter plots of GECQE and LCP scores.}
    \label{fig:scatter_0-9}
\end{figure}

\end{document}